\pdfoutput=1
\documentclass[11pt]{article}

\PassOptionsToPackage{table,xcdraw}{xcolor}
\usepackage[final]{acl}
\usepackage{multirow}
\usepackage{times}
\usepackage{titlesec}
\usepackage{latexsym}
\usepackage{float}
\usepackage{makecell}
\usepackage{enumitem}
\usepackage{graphicx}
\usepackage{booktabs}
\usepackage{array}
\usepackage{longtable}
\usepackage{lipsum}

\usepackage{booktabs}
\usepackage{tabularx}
\usepackage{amsmath}
\usepackage{amssymb}
\usepackage[utf8]{inputenc}

\usepackage{array}
\usepackage{multirow}
\usepackage{graphicx}
\usepackage{longtable}
\usepackage{hyperref}
\usepackage{times}
\usepackage{svg}
\usepackage{latexsym}
\usepackage{multirow}
\usepackage{threeparttable}
\usepackage[ruled,vlined,linesnumbered]{algorithm2e}
\usepackage{amsmath, amssymb} % 数学符号支持
\usepackage{amsmath}
\usepackage{graphicx}
\usepackage{booktabs}
\usepackage{adjustbox}

\usepackage{listings}
\usepackage[T1]{fontenc}
\usepackage[utf8]{inputenc}
\usepackage{multirow}
\usepackage{microtype}

\usepackage{tikz}
\newcommand*\circled[1]{\tikz[baseline=(char.base)]{\node[shape=circle,fill=black,text=white,draw,inner sep=.1pt] (char) {#1};}}

\setlist{nosep}

\usepackage{amsmath}
\title{GRAIN: Bridging Name and Narrative Shifts in Real-World Graph Reasoning through Invariance-Rewarded Agentic RL}

\author{
    \textbf{Zike Yuan\textsuperscript{1,2}},
    \textbf{Han Zhang\textsuperscript{2}},
    \textbf{Jianzhi Yan\textsuperscript{1,2}},
    \textbf{Le Liu\textsuperscript{1,2}},
    \textbf{Cai Ke\textsuperscript{1,2}},
    \textbf{Huozhi Zhou\textsuperscript},
    \textbf{Jian Xie\textsuperscript},\\
    \textbf{Jiran Yin\textsuperscript},
    \textbf{Yukun Cao\textsuperscript{3}},
    \textbf{Yue Yu\textsuperscript{2}},
    \textbf{Hui Wang\textsuperscript{2}},
    \textbf{Ming Liu\textsuperscript{1,2,*}},
    \textbf{Bing Qin\textsuperscript{1,2,*}}
    \\
    \textsuperscript{1}Harbin Institute of Technology, Shenzhen, China\\
    \textsuperscript{2}Peng Cheng Laboratory, Shenzhen, China\\
    \textsuperscript{3}Xidian University, Xi'an, China\\
    \texttt{\{yuanzk,wangh06\}@pcl.ac.cn}\\
    \texttt{\{mliu,qinb\}@ir.hit.edu.cn}\\
    \texttt{\ caoyukun@xidian.edu.cn}
}

\begin{document}
\maketitle
\begingroup
\renewcommand{\thefootnote}{*}
\footnotetext{\scriptsize Corresponding authors:  Ming Liu, and Bing Qin.}
\endgroup
\enlargethispage{\baselineskip}
\begin{abstract}
Despite their potential in standardized graph tasks, Large Language Models (LLMs) remain brittle to real-world shifts in node identifiers and task formulation. While deterministic graph tools are invariant to such shifts, extracting topological structures from noisy text is highly fragile for LLMs, which often overfit to surface patterns. Moreover, mitigating these parsing failures via multi-agent systems incurs prohibitive latency. To address this, we propose GRAIN, a single-agent framework optimized via reinforcement learning. GRAIN models reasoning as a semantic parsing and tool-execution pipeline, guided by a Structure Invariance Reward. By validating extracted intermediate graphs against ground-truth topologies, this reward forces the LLM to learn robust text-to-structure mappings rather than memorizing linguistic artifacts. We also introduce GRIT, a benchmark evaluating sensitivity to such linguistic shifts. GRAIN outperforms multi-agent baselines by 16.45\% in accuracy with approximately 24\% lower latency. Furthermore, it demonstrates superior structural generalization, halving the out-of-distribution (OOD) gap of SFT models (from 15.77\% to 7.80\%) and maintaining robustness on large-scale graphs beyond the training distribution.
\end{abstract}

\section{Introduction}

\begin{figure}[t]
  \centering
  \includegraphics[width=0.5\textwidth]{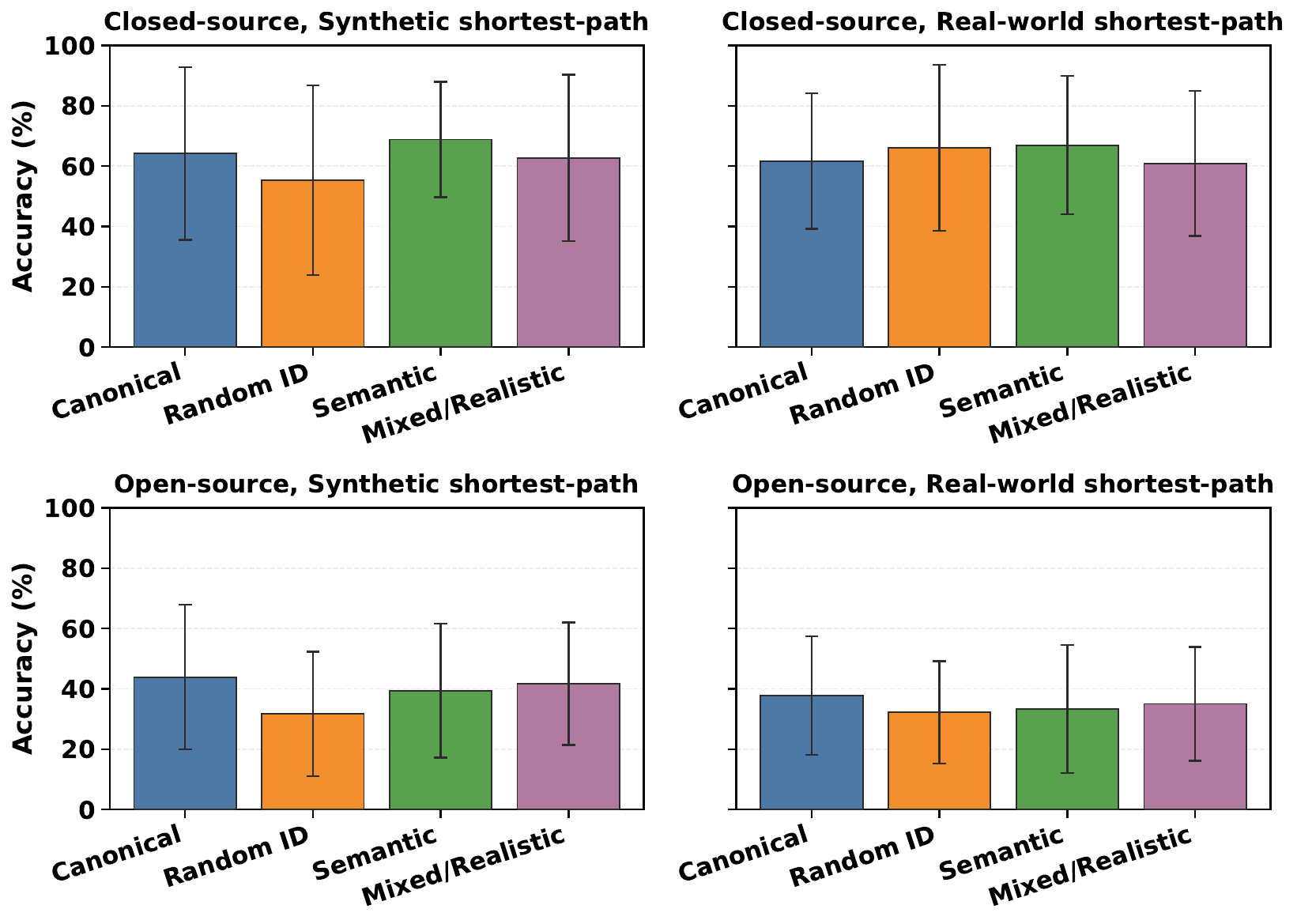}
\caption{\textbf{Node-label Sensitivity.} Large error bars reveal high instability: open-source models degrade on Random IDs, while closed-source models struggle with canonical forms.}
  \label{fig:2} 
       \vspace{-20pt}
\end{figure}

Large Language Models (LLMs) exhibit great versatility in language, code, and tool utilization. However, critical domains including social network analysis, knowledge graph reasoning, and software dependency management necessitate robust reasoning over inherent graph structures. While recent approaches address classical tasks using text-serialized graphs~\citep{NLgraph-6}, they often depend on benchmarks with sanitized identifiers. However, extracting accurate graph structures from text heavily laden with naming irregularities, alias collisions, and semantic noise is a fundamental challenge in Information Extraction (IE) and Semantic Parsing. Such idealized settings overlook these real-world linguistic complexities, which ultimately limits reliability in practical applications.
\begin{figure}[t]
  \includegraphics[width=\columnwidth]{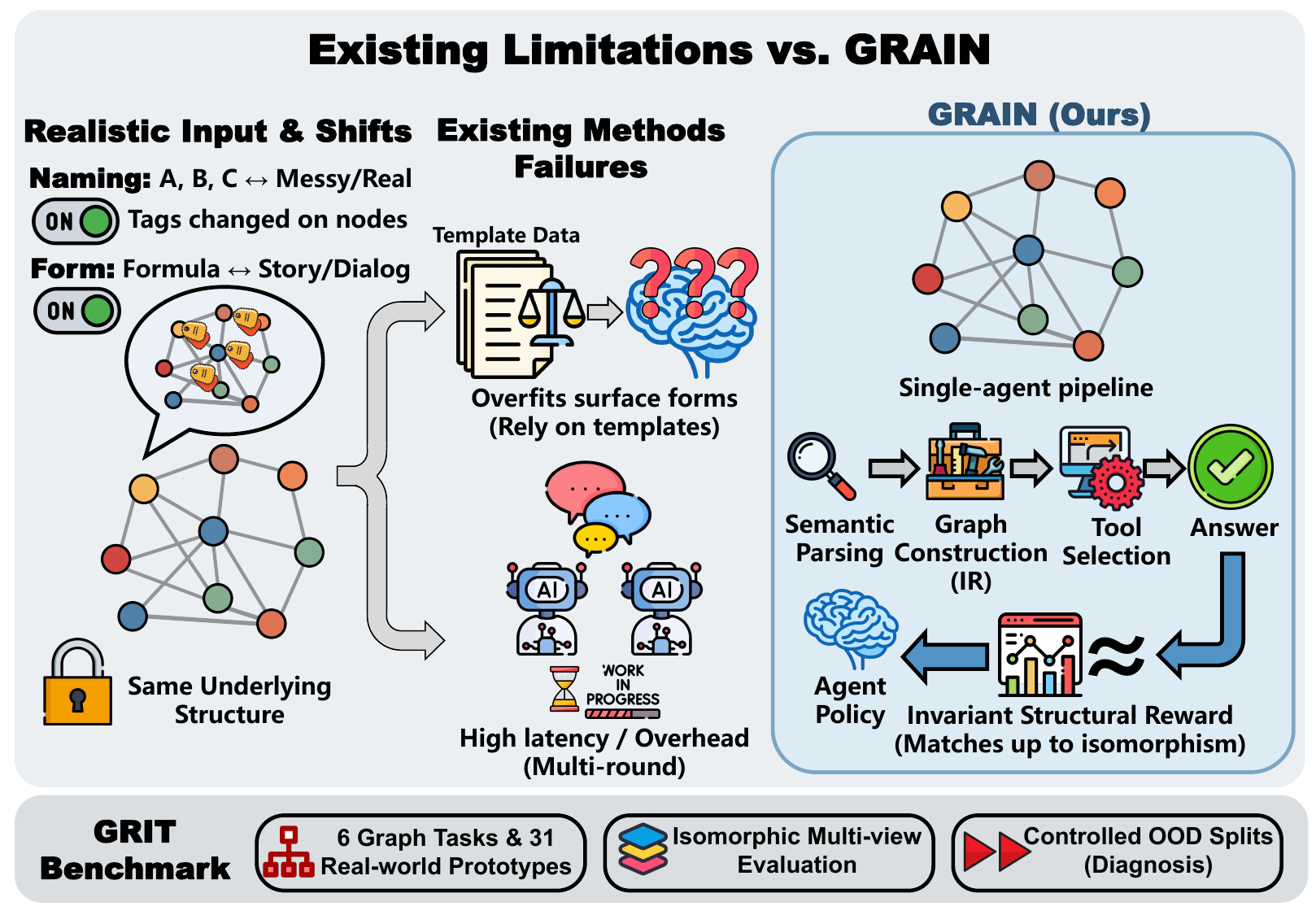}
  \vspace{-20pt}
\caption{GRAIN mitigates sensitivity to \textbf{naming and formulation shifts} in real-world graph reasoning via a single-agent tool-use pipeline trained with \textbf{isomorphism-aware structural rewards}.}

  \label{fig:1}
\end{figure}
We examine the robustness gap between clean benchmarks and practical graph reasoning tasks, revealing that LLMs are surprisingly brittle to superficial variations. As shown on the left side of Figure~\ref{fig:1}, this instability stems from two sources: \circled{1} \textbf{Node-label sensitivity}, where altering naming schemes (e.g., Random vs. Semantic) shifts performance despite structural equivalence. \circled{2} \textbf{Task-form sensitivity}, where performance degrades when moving from synthetic formulations to real-world narratives. Our analysis (see Figure~\ref{fig:2} and Table~\ref{tab:model_performance}) highlights a distinct disparity: while closed-source models generally maintain accuracy but suffer from unpredictable variance across naming schemes, open-source models exhibit severe degradation under non-canonical settings, especially in realistic scenarios. These observations underscore the critical need to explicitly model and mitigate robustness failures caused by naming and formulation shifts.

Prior research investigates LLM graph representations, yet critical limitations persist. \textbf{First}, while textualization studies~\citep{talkgraph-9} reveal sensitivity to serialization (e.g., labeling), they typically overlook distribution shifts between standardized templates and noisy real-world formulations. \textbf{Second}, optimization paradigms like CoT and Supervised Fine-Tuning (SFT) struggle with structural generalization; SFT specifically overfits surface patterns, causing performance degradation under isomorphic variations, scale shifts (\textit{i.e.}, generalizing to graphs with significantly more nodes than the training set), or computational intensity (e.g., NP-hard problems)~\citep{chen2024graphwiz,guo2025g1,zhang2025generalizable}. \textbf{Finally}, although multi-agent frameworks~\citep{yuan2025ma,zhang2024gcoder} improve accuracy via external tools, they suffer from inherent inefficiency: reliance on multi-round interactions incurs prohibitive token costs and latency, rendering them impractical for real-time deployment.

To address these limitations, we identify three critical design requirements:
(i) \textbf{Explicit Pipeline Modelling}: Instead of relying on standard, free-form Chain-of-Thought (CoT) generation where LLMs mix logical reasoning with mental arithmetic, we decompose reasoning into explicit \textit{semantic parsing} and \textit{algorithmic execution}. This imposes verifiable constraints on intermediate representations, ensuring structural validity before computation.
(ii) \textbf{Structural Invariance}: To decouple reasoning from specific identifiers, we vary naming and narratives while preserving the underlying structure, compelling the model to learn invariant rules rather than memorizing surface artifacts.
(iii) \textbf{Unified Agentic RL Optimization}: As static supervision overfits surface forms, we employ RL with structural rewards to enforce generalization. This agentic RL formulation distills the high-accuracy reasoning capabilities---typically requiring complex multi-agent collaboration---into a single model's weights, achieving state-of-the-art performance without incurring their prohibitive latency.

Guided by these principles, we propose \textbf{GRAIN} (\textbf{G}raph \textbf{R}easoning \textbf{A}gent with \textbf{IN}variance), a single-agent RL framework for robust graph reasoning. GRAIN acts as an agentic pipeline that autonomously parses noisy queries, selects optimal external algorithms, and constructs necessary structural arguments. This ensures computational correctness while focusing the LLM purely on semantic parsing and information extraction. To mitigate sensitivity to surface shifts, we train GRAIN on diverse isomorphic narratives. Crucially, we optimize a unified RL objective with an \textit{invariance-oriented structural reward}. By validating generated intermediate graphs against ground-truth topologies, this signal provides direct feedback on structure recovery, forcing the policy to learn invariant structural rules rather than memorizing superficial identifiers.

To facilitate systematic diagnosis, we introduce \textbf{GRIT}, a multi-task corpus covering six graph problems across 31 real-world prototypes. By pairing standard formulations with diverse narratives and namings, GRIT enables controlled \textbf{OOD} evaluation on fixed structures. Experiments demonstrate that GRAIN leverages this structural feedback to optimize the ``language $\to$ structure'' chain, achieving exceptional robustness against surface shifts while maintaining low computational complexity. Our primary contributions are:

\begin{itemize}[left=0pt]
\item We quantify LLM sensitivity to node labeling and task formulation under a controlled isomorphic setup. We demonstrate that surface-level variations induce significant performance volatility, even when the underlying structure and semantics remain fixed.
\item We propose \textbf{GRAIN}, a single-agent RL framework that leverages an invariance-oriented structural reward. By directly optimizing intermediate structure recovery and tool invocation, GRAIN achieves superior robustness across diverse naming schemes and task forms.
\item We release \textbf{GRIT}, a multi-task benchmark featuring isomorphic multi-view coverage with explicit OOD splits. This resource enables systematic, reproducible evaluation of graph reasoning robustness against naming and formulation shifts.
\end{itemize}

\section{Related Work}

\paragraph{LLMs for Graph Reasoning.}
Graph reasoning is essential to many real-world applications, including social network analysis, knowledge graph reasoning ~\citep{chen2024large,li2026kg}, and software dependency management. However, existing benchmarks show that LLM performance declines substantially as graph size and task complexity increase, particularly when moving from basic connectivity problems to NP-hard tasks~\citep{NLgraph-6,tang2024grapharena,yuan2025gracore,luo2024graphinstruct,Zhang2024CanLG}. To better understand these limitations, recent research has begun to shift from evaluating only final answers toward diagnosing intermediate reasoning processes~\citep{taylor2024large}. Meanwhile, instruction-tuning methods improve graph reasoning accuracy through specialized data and curricula~\citep{cao2024graphinsight,chen2024graphwiz,guo2025g1,wang2025graph,yuan2026egl}. Nevertheless, these methods are typically optimized for standardized inputs and therefore remain brittle to changes in graph serialization formats and prompting schemes~\citep{xu2025graphomni}. This gap highlights the need for LLMs that can reason robustly over inherent graph structures rather than relying on specific surface representations.

\paragraph{Serialization Sensitivity.}
Graph serialization is a critical determinant of LLM performance. Prior work reveals that different encoding schemes~\citep{talkgraph-9} and even the descriptive order of edges~\citep{li2024can} can drastically alter reasoning accuracy, highlighting a severe sensitivity to surface-form variations. To mitigate this reliance on raw textualization, approaches like GraphToken~\citep{perozzi2024let} propose injecting structured signals via parameter-efficient encodings, though robustness against diverse naming conventions remains an open challenge.

\paragraph{Agents and Tool Learning.}
Tool-augmented paradigms enhance reliability via external APIs or code execution~\citep{yao2022react,schick2023toolformer,qin2023toolllm,jin2025search,zhang2024effective,zeng2025autogen,xiong2026adaptive}. In the graph domain, multi-agent frameworks improve performance by decomposing tasks into coordinated roles~\citep{li2024graphteam,yuan2025ma,han2025see,cao2024lego,qian2025toolrl}. Yet, these systems incur high latency and token costs due to extensive inter-agent communication, highlighting the necessity for efficient single-agent solutions.

\begin{figure*}[t]
  \centering
  \includegraphics[width=\textwidth]{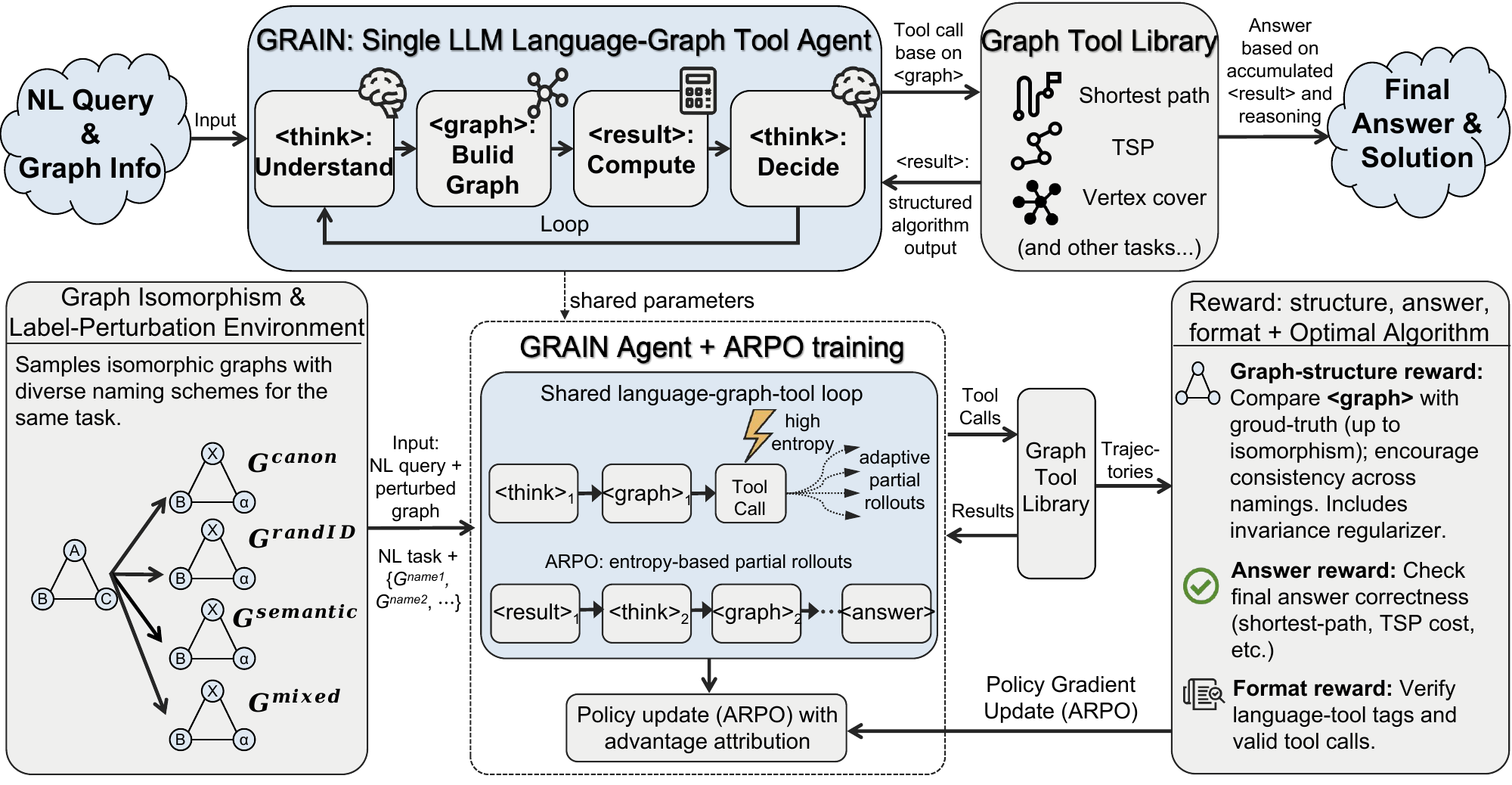} % 请替换为您的图片路径
  \caption{\textbf{Overview of the GRAIN framework.} 
\textbf{(Top) Inference:} The LLM builds a graph IR in \texttt{<graph>}, calls solvers, and outputs the answer. 
\textbf{(Bottom) Training:} \textbf{ARPO} trains under isomorphism-preserving name perturbations, with partial rollouts and a reward for structural invariance, correctness, and valid format.
}
  \label{fig:3}

\end{figure*}

\section{GRIT: A Benchmark for Robust Graph Reasoning}
\label{sec:grit}

We introduce GRIT to evaluate LLM robustness against two real-world distribution shifts: \textit{identifier shift} (node naming variations) and \textit{task-form shift} (standard vs. narrative formulations). Unlike benchmarks with fixed identifiers, GRIT provides controlled multi-view instances varying surface realizations while preserving underlying structures, serving as a rigorous diagnostic tool for structural information extraction and semantic parsing.

\begin{table}[t]
\begin{center}
\small 
\begin{tabular*}{\columnwidth}{@{\extracolsep{\fill}}lccccc}
\toprule
\textbf{Split} & \makecell[c]{\textbf{Node}\\\textbf{Range}} & \makecell[c]{\textbf{\#}\\\textbf{Tasks}} & \makecell[c]{\textbf{\#}\\\textbf{Scen.}} & \makecell[c]{\textbf{\#}\\\textbf{Graphs}} & \makecell[c]{\textbf{Total}\\\textbf{Qs.}} \\
\midrule
Train & 4--40 & 6 & 31 & 2160 & 17,280 \\
Test & 10--40 & 6 & 31 & 360 & 2,760 \\
Large Test & 40--60 & 6 & 31 & 120 & 960 \\
OOD Test & 10--40 & 6 & 24\textsuperscript{*} & 180 & 1,440 \\
\bottomrule
\multicolumn{6}{c}{\footnotesize \textit{\textsuperscript{*}Denotes distinct scenarios unseen during training.}} \\
\end{tabular*}
\caption{Statistics of the GRIT benchmark splits.}
\label{tab:1}
\end{center}
          \vspace{-25pt}
\end{table}

\paragraph{Task coverage and scenarios.}
GRIT covers six fundamental tasks spanning polynomial to NP-hard complexities~\citep{zhang2006introduction}. We instantiate them into 31 scenarios (e.g., logistics routing, social networks), yielding naturalistic narratives laden with complex syntactic embeddings and semantic distractors. This offers significantly richer linguistic variation than canonical definitions (details in Appendix~\ref{sec:app_grit}).

\paragraph{Multi-view question construction.}
Each base instance generates \textbf{8} distinct queries by crossing \textbf{two forms} (\emph{Standard} vs. \emph{Realistic} narrative) with \textbf{four naming schemes}: \textit{Canonical}, \textit{Random IDs} (testing tokenization robustness), \textit{Semantic}, and \textit{Noisy/Mixed} (testing coreference resolution and alias mapping). This factorial design isolates specific axes of variation, enabling controlled evaluation of both identifier and task-form shifts.

\paragraph{Graph generation and deterministic injection.}
GRIT features topologies spanning in-domain ($N \in [10, 40]$) and length-generalization ($N \in [40, 60]$) scales, with ground-truth answers derived via symbolic solvers. Crucially, to prevent hallucinations, narratives are \textit{not} LLM-generated. Instead, we employ a \textbf{strictly deterministic, template-based slot-filling engine} to inject topologies into linguistic scenarios, guaranteeing 100\% reliable isomorphism between the text and the underlying graph.

\paragraph{Benchmark splits.}
Table~\ref{tab:1} summarizes the dataset splits. \textbf{Train} and \textbf{Test} share the same scenario pool, while \textbf{Large Test} evaluates length generalization on larger graphs ($N \in [40,60]$). To isolate robustness against surface-level variations, the \textbf{OOD Test} set introduces \emph{unseen} held-out scenarios and identifier realizations not observed during training.

\section{GRAIN}
\label{sec:method}

To address the brittleness caused by identifier and task shifts, we propose \textbf{GRAIN}, a single-agent reinforcement learning framework for robust graph reasoning. GRAIN models the LLM as a language--graph--tool agent trained in an isomorphism-based environment with diverse naming and formulation views. Crucially, we enforce a \textit{gated} reward structure that strictly penalizes format violations while incentivizing recovery of the underlying topology, ensuring the policy captures structural rules invariant to surface-level labels and wording.

\subsection{Problem Formulation and Overall Objective}
\label{sec:problem}
To formalize the task (Table~\ref{tab:notations_unified}), let $G=(V,E,w)$ be a weighted graph and $\mathcal{T}$ a task (e.g., TSP). Real-world queries incorporate variations in node naming $\nu:V\rightarrow\Sigma^{*}$ and formulation style $\phi$ (e.g., narrative complexity), defined as the natural language query $x = f(G,\nu,\phi,\mathcal{T})$. We denote the training graph distribution by $\mathcal{G}$ and the conditional distribution over surface forms $(\nu,\phi)$ given $G$ by $\mathcal{P}(G)$.

Given a policy $\pi_{\theta}$, the full generation for input $x$ is a trajectory $\tau(G,\nu,\phi;\pi_{\theta})$, containing all tokens from \texttt{<think>} to \texttt{</answer>}. At a high level, we aim to train a policy that maximizes task performance while simultaneously recovering the underlying graph structure, robust to variations in naming and phrasing. We formulate this as maximizing the expected joint return:
\begin{equation}
\label{eq:high_level_objective}
\max_{\theta}\;
\mathbb{E}_{\substack{G\sim\mathcal{G} \\ (\nu,\phi)\sim\mathcal{P}(G)}}
\Big[
  R\big(\tau(G,\nu,\phi;\pi_{\theta})\big)
\Big],
\end{equation}
where the total return $R(\tau)$ composites the task-level reward and a structural invariance score, subject to strict formatting constraints as detailed in Section~\ref{sec:env_reward}.

\subsection{GRAIN: A Single Language--Graph--Tool Agent}
\label{sec:agent}

GRAIN treats the LLM as a single agent interacting with a graph-tool library via structured text. For each input $x$, the agent outputs a sequence with four tagged segments (a complete step-by-step execution example is provided in Appendix~\ref{sec:app_case_study}):
\begin{equation}
\begin{split}
y = \big\langle & \texttt{<think>}\dots, \texttt{<graph>}\dots, \\
& \texttt{<result>}\dots, \texttt{<answer>}\dots \big\rangle.
\end{split}
\end{equation}

As illustrated in Figure~\ref{fig:3}, the overall reasoning pipeline consists of four stages:

\paragraph{Task understanding and planning (\texttt{<think>}).}
The agent reads $x$, identifies the task type $\mathcal{T}$, extracts entity mentions corresponding to nodes and edges, and sketches a plan of which graph algorithms to call and with which parameters.

\paragraph{Graph construction (\texttt{<graph>}).}
In the \texttt{<graph>} segment, the agent emits a structured representation $\hat{G}$ that specifies the node set, edges and weights, and task-specific parameters (e.g., source/target nodes or the set of cities for TSP). The runtime system parses \texttt{<graph>} into the input format of the graph-tool library.

\paragraph{Tool calls and result injection (\texttt{<result>}).}
The system invokes the corresponding graph algorithm, obtains a deterministic result $\hat{z}$, and injects it back into the context as a \texttt{<result>} segment.

\paragraph{Continued reasoning and final answer (\texttt{<answer>}).}
Upon seeing \texttt{<result>}, the agent may produce additional \texttt{<think>} and \texttt{<graph>} segments and trigger more tool calls, or directly synthesize the final answer in \texttt{<answer>}. The episode terminates at \texttt{</answer>}.

With the tool library ensuring computational correctness, learning focuses on the language-to-graph decision chain: entity resolution, reconstruction, and tool/parameter selection under noisy naming and diverse formulations.

\subsection{Isomorphic Environment and Gated Rewards}
\label{sec:env_reward}

To simulate real-world variability, we construct an isomorphism-based perturbation environment. For each training graph $G$ and task $\mathcal{T}$, we generate diverse naming schemes $\nu$ (Canonical, Random, Semantic, Mixed) and query styles $\phi$, yielding an instance family $x = f(G,\nu,\phi,\mathcal{T})$.

The RL state $s_t$ comprises the input $x$, generated history, and injected \texttt{<result>} segments; episodes terminate at \texttt{</answer>}. To enforce strict protocol adherence while optimizing structural recovery, we employ a \textbf{gated reward formulation} consisting of three components:

\paragraph{Answer Reward $r_{\text{ans}}(\tau)$.}
This term checks whether the \texttt{<answer>} segment provides the correct task-specific value (shortest-path length, TSP tour cost, etc.) and assigns a binary or shaped reward accordingly.

\definecolor{grainhighlight}{HTML}{F0F6F6}
\begin{table*}[t]
\centering
\small 
% 1. 保持行高，舒展
\renewcommand{\arraystretch}{1.3} 
% 2. 将列间距设为0，由 extracolsep 自动计算填充
\setlength{\tabcolsep}{0pt} 

% 3. 使用 tabular* 并设置宽度为 \textwidth
% 4. @{\extracolsep{\fill}} 是关键，它会自动拉伸列间距
\begin{tabular*}{\textwidth}{@{\extracolsep{\fill}} ll cccccc c }
\toprule
\textbf{Model Family} & \textbf{Method / Setting} & \textbf{S. Path} & \textbf{Coloring} & \textbf{TSP} & \textbf{V. Cover} & \textbf{BFS} & \textbf{Centr.} & \textbf{Avg.} \\ 
\midrule

% --- Group 1 ---
\multicolumn{9}{l}{\textit{\textbf{Proprietary Models \& Multi-Agent Systems}}} \\
\multirow{2}{*}{GPT-5-nano} & Zero-shot CoT & 72.18 & 43.54 & 3.33 & 29.17 & 16.25 & 15.21 & 29.95 \\
                            & Tool-use CoT  & 73.39 & 68.96 & 41.88 & 28.61 & 31.46 & 40.42 & 47.45 \\
\cmidrule(l){2-9}
G1-3B                       & Zero-shot CoT & 11.46 & 8.96 & 0.00 & 5.28 & 1.04 & 8.33 & 5.85 \\
MA-GTS(GPT-4o-mini)                      & Multi-Agent   & 85.63 & 87.50 & 91.25 & 68.33 & 55.62 & \underline{87.50} & 79.31 \\
MA-GTS(Qwen3-4B-Ins.)$^\dagger$         & Multi-Agent   & 13.75 & 12.50 & 1.88 & 4.17 & 0.62 & 7.50 & 6.74 \\
\midrule

% --- Group 2 ---
\multicolumn{9}{l}{\textit{\textbf{Llama-3.2 Series (3B)}}} \\
\multirow{2}{*}{Llama-3.2-Ins} & Zero-shot CoT & 1.88 & 13.96 & 0.21 & 3.61 & 4.38 & 4.58 & 4.77 \\
                               & Tool-use CoT  & 2.92 & 11.88 & 0.21 & 4.44 & 0.83 & 5.42 & 4.28 \\
\rowcolor{grainhighlight}
\textbf{GRAIN (Ours)}          & SFT + ARPO    & 45.93 & 51.46 & 34.79 & 15.83 & 71.88 & 64.58 & 47.41 \\
\midrule

% --- Group 3 ---
\multicolumn{9}{l}{\textit{\textbf{Qwen-3 Series (4B)}}} \\
\multirow{2}{*}{Qwen-3-Ins}    & Zero-shot CoT & 53.13 & 28.75 & 1.88 & 5.28 & 70.42 & 12.71 & 28.69 \\
                               & Tool-use CoT  & 48.13 & 36.88 & 71.25 & 38.61 & 60.42 & 42.29 & 49.60 \\
\cmidrule(l){2-9} 
\multirow{4}{*}{Qwen-3-Base}   & Zero-shot CoT & 49.38 & 22.29 & 2.08 & 4.44 & 48.96 & 11.88 & 23.17 \\
                               & Tool-use CoT  & 35.42 & 45.28 & 40.00 & 37.50 & 70.42 & 42.29 & 45.15 \\
                               & SFT (Text-only) & 22.13 & 46.88 & 3.54 & 33.89 & 71.25 & 45.00 & 37.11 \\
                               & SFT (Tool-use CoT) & \textbf{99.58} & \textbf{93.13} & \underline{96.88} & \underline{88.33} & \underline{82.50} & 74.17 & \underline{89.10} \\
\rowcolor{grainhighlight} 
\textbf{GRAIN (Ours)}          & SFT + ARPO    & \underline{99.38} & \underline{92.92} & \textbf{97.92} & \textbf{93.30} & \textbf{98.33} & \textbf{92.70} & \textbf{95.76} \\ 
\bottomrule
\end{tabular*}
\caption{\textbf{Main Results on GRIT Tasks.} We report the accuracy (\%) across six graph reasoning tasks. The best results are \textbf{bolded}, and the second-best results are \underline{underlined}. ``Ins'' denotes Instruct models. For Qwen-3-Base, we compare CoT and SFT under both text-only and tool-use settings. $^\dagger$ denotes zero-shot transfer of the six-agent MA-GTS pipeline to Qwen3-4B-Instruct on the 920-example Test set; its micro-average is 6.85\%. GRAIN consistently achieves SOTA performance.}
\label{tab:2}

\end{table*}

\paragraph{Structure Similarity Score $s_{\text{inv}}(\tau)$.}
We parse the extracted graph $\hat{G}$ from the \texttt{<graph>} segment and compute the \textbf{Jaccard Index} over the canonical edge sets of $\hat{G}$ and the ground-truth $G$. Specifically, we align entity names via the environment's mapping and represent edges as sets of canonical tuples \texttt{(source, target, weight)}. This set-theoretic implementation mathematically prevents out-of-bounds dimension errors that would occur in matrix-based similarity (e.g., Cosine) if the LLM hallucinates or misses nodes. Furthermore, unlike computationally NP-hard metrics such as Graph Edit Distance (GED), this signal is highly efficient to compute while granting dense, continuous partial credit ($s_{\text{inv}} \in [0, 1]$) for topological recovery.

\paragraph{Gated Total Reward $R(\tau)$.}
Instead of treating formatting as a soft regularization term, we impose it as a hard constraint. The final trajectory return is defined as:
\begin{equation}
\label{eq:gated_reward}
R(\tau) = 
\begin{cases} 
\rho_{\text{err}}, & \text{if invalid}, \\
r_{\text{ans}}(\tau) + \lambda_{\text{inv}}\,s_{\text{inv}}(\tau), & \text{if valid}.
\end{cases}
\end{equation}
where $\rho_{\text{err}}$ penalizes malformed trajectories (e.g., missing tags). For valid outputs, the reward sums task correctness and structure similarity (equally weighted with $\lambda_{\text{inv}}=1.0$).

Training utilizes the underlying $G$ to compute $s_{\text{inv}}(\tau)$, whereas inference operates without labels. By scoring diverse views $(\nu,\phi)$ of the same $G$ with this signal, the policy learns invariant topological rules rather than memorizing surface phrasing.

\subsection{Reinforcement Learning with ARPO}
\label{sec:arpo}

To address cold-start challenges and ensure format compliance, we perform a brief \textbf{SFT warm start}. Expert trajectories are automatically generated on small synthetic graphs ($|V|\le 14$) using the graph-tool library. Minimizing standard autoregressive cross-entropy loss enables the model to acquire correct tag usage, valid tool syntax, and basic construction skills. This model serves as both the RL initialization $\theta_0$ and the reference policy $\pi_{\text{ref}}$.

We then further optimize the policy using Reinforcement Learning with Verifiable Rewards (RLVR). Writing $x=f(G,\nu,\phi,\mathcal{T})$ and $\tau\sim\pi_{\theta}(\cdot\mid x)$, our training objective is:
\begin{equation}
\label{eq:rlvr_objective}
\begin{split}
J(\theta) &= \mathbb{E}_{x,\,\tau\sim\pi_{\theta}(\cdot\mid x)} \Big[ R(\tau) \\
  &\quad - \beta \,\mathrm{KL}\big(
      \pi_{\theta}(\cdot\mid x)
      \,\|\,
      \pi_{\text{ref}}(\cdot\mid x)
    \big)
\Big],
\end{split}
\end{equation}
where $R(\tau)$ follows the gated formulation in Eq.~\eqref{eq:gated_reward}. We optimize $J(\theta)$ using \textbf{ARPO} (Agentic Reinforced Policy Optimization) \citep{dong2025arpo}, which is uniquely suited for our problem setting:

\paragraph{Entropy-based adaptive partial rollouts.}
In our pipeline, naming and query perturbations primarily impact entity resolution and graph reconstruction, empirically manifesting as token entropy spikes. ARPO detects these high-uncertainty moments to trigger adaptive partial rollouts (branching), enabling the agent to explore diverse graph construction hypotheses when confronting ambiguous identifiers.

\paragraph{Branch-aware advantage estimation.}
ARPO standardizes returns across trajectories sharing a prefix, assigning shared advantages to prefix tokens (task understanding) and individual ones to branch suffixes (graph construction). This ensures the structure reward $s_{\text{inv}}$ targets high-entropy grounding decisions, facilitating efficient learning.

\section{Evaluation}

\subsection{Experimental Setup}
\paragraph{Benchmarks and Data.}
We evaluate primarily on \textbf{GRIT}, spanning six tasks (\textsc{Shortest Path}, \textsc{Coloring}, \textsc{TSP}, \textsc{Vertex Cover}, \textsc{BFS}, \textsc{Centrality}). Each instance features \textbf{8 views} derived from \textbf{31} scenarios: \textbf{2} forms (\textit{Standard}/\textit{Realistic}) $\times$ \textbf{4} naming schemes (\textit{Canonical}, \textit{Random IDs}, \textit{Semantic}, \textit{Noisy/Mixed}). Training uses graphs with $N\le 40$; evaluation covers \textbf{ID} (seen), \textbf{OOD-Form} (unseen naming--narrative pairs, $N\le 40$), and \textbf{OOD-Large-Size} ($N \in [40, 60]$). To ensure unbiased evaluation, the node sizes in our test sets follow a strictly balanced, uniform distribution (detailed in Appendix~\ref{sec:appendix_size_dist}). We also report results on external benchmarks \textbf{GraphInstruct}~\citep{luo2024graphinstruct} and \textbf{GraphArena}~\citep{tang2024grapharena}.

\paragraph{Models and Baselines.}
We compare \textbf{Zero-shot CoT}~\citep{wei2022chain}, \textbf{Tool-use CoT}, \textbf{SFT}, and \textbf{GRAIN} (SFT+\textbf{ARPO}). Baselines include \textbf{GPT-5-nano}~\citep{gpt4}, the graph-specific RL model \textbf{G1-3B}~\citep{guo2025g1}, and the multi-agent \textbf{MA-GTS}~\citep{yuan2025ma}. For open models, we evaluate \textbf{Llama-3.2-Ins}~\citep{llama} (3B) and \textbf{Qwen-3-Ins}~\citep{qwen} (4B) (``Ins'' denotes Instruct). Additionally, we train \textbf{GRAIN} on \textbf{Llama-3.2} (3B) to test transferability. On \textbf{Qwen-3-Base} (4B), we conduct a comprehensive comparison across Zero-shot, Tool-use, SFT (Text/Tool), and GRAIN to isolate the specific effects of tool access and training.

\begin{figure}[t]
\vspace{-5pt}
\centering
  \includegraphics[width=0.5\textwidth]{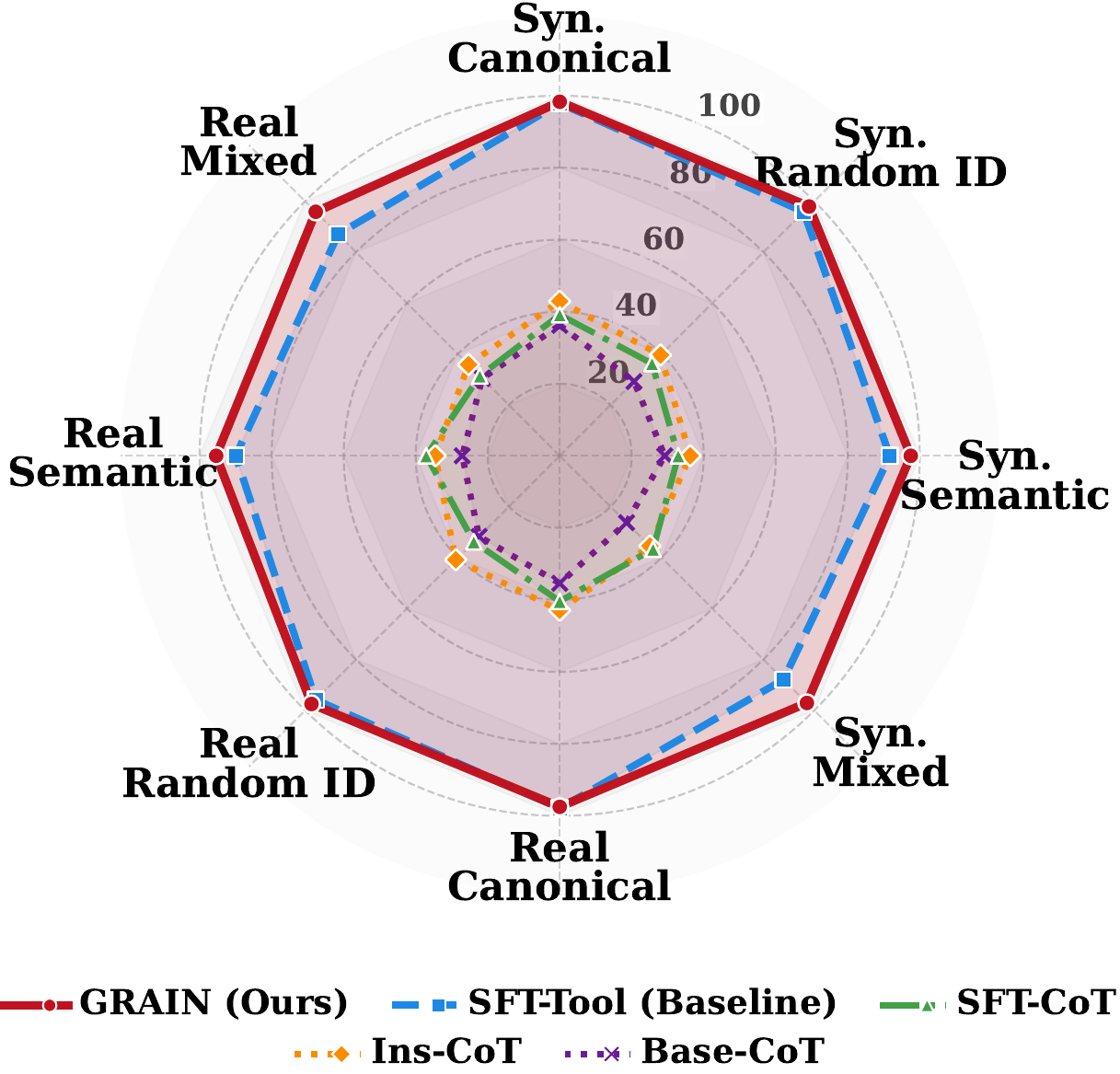}
  \vspace{-20pt}
\caption{
\textbf{Robustness Evaluation.} GRAIN (Red) achieves consistent robustness across shifts, unlike SFT-Tool (Blue) which degrades significantly in noisy, real-world scenarios.}
    \label{fig:4} 
            \vspace{-15pt}
\end{figure}

\paragraph{Tool Library and Inference Protocol.}
Tool-use methods share a unified interface: models generate graph representations, invoke verifiable solvers, and derive answers. We enforce identical decoding budgets and tool-call limits (text-only baselines disable tools).

\paragraph{Training Details.}
Training relies exclusively on \textbf{GRIT Train}. \textbf{SFT} is performed on the full split to provide initializations. \textbf{GRAIN} warm-starts from the tool-use SFT checkpoint, employing a two-stage curriculum: first stabilizing structured generation on small graphs (\textbf{4--14} nodes), then running \textbf{ARPO} on the \textbf{10--40} node range by sampling \textbf{two} underlying graphs per task and size. Implementation details are provided in Appendix~\ref{sec:app_training}.

\paragraph{Evaluation Metrics.}
We report \textbf{final-answer accuracy} (\%), counting a prediction as correct iff it strictly matches the ground truth. Tables present per-task performance and the macro-average (\textbf{Avg.}), computed using a unified scoring script across all settings.

\subsection{Performance}

\subsubsection{Overall Performance on GRIT}
\label{sec:exp_overall}

Table~\ref{tab:2} compares GRAIN against representative baselines across six graph reasoning tasks. GRAIN achieves consistent state-of-the-art performance across varying model scales, demonstrating the effectiveness of our proposed framework. The analysis of the case study is presented in the Appendix~\ref{sec:app_case_study}.

\paragraph{Surpassing Supervised Baselines.} 
GRAIN significantly outperforms SFT under identical configurations. On Qwen-3-Base, GRAIN boosts the strong SFT baseline (89.10\%) to \textbf{95.76\%}. Substantial gains on complex tasks like Vertex Cover (+4.97\%) and BFS (+15.83\%) validate that our \textit{invariance-oriented structural reward} effectively optimizes non-differentiable decision chains, surpassing the performance ceiling of standard imitation learning.

\paragraph{Multi-Agent Comparison.}
MA-GTS with GPT-4o-mini achieves 79.31\%, whereas GRAIN reaches 95.76\% with a compact Qwen-3 backbone. As a compact-model transfer diagnostic, the same six-agent MA-GTS pipeline with Qwen3-4B-Instruct obtains a 6.74\% macro-average (6.85\% micro-average) on the 920-example Test set, with consistently low accuracy across tasks. Because training, backbone initialization, and agent organization differ from GRAIN, this result should not be interpreted as an isolated single- versus multi-agent effect; rather, it indicates that inference-only multi-agent decomposition does not automatically resolve graph-grounding and executable tool-call failures in this setting. Appendix~\ref{sec:appendix_failure_analysis} reports the complete pipeline diagnostic.

\definecolor{grainhighlight}{HTML}{F0F6F6}
\begin{table}[t]
\centering
\small
\renewcommand{\arraystretch}{1.25} 
\setlength{\tabcolsep}{0pt}      
\begin{tabular*}{\columnwidth}{@{\extracolsep{\fill}} l ccccc }
\toprule
\multicolumn{1}{c}{\multirow{2}{*}{\textbf{Method}}} & 
\multicolumn{3}{c}{\textbf{GRIT Distribution}} & 
\multicolumn{2}{c}{\textbf{Unseen Tasks}} \\
\cmidrule(lr){2-4} \cmidrule(l){5-6}
 & \textbf{Test} & \textbf{OOD} & \textbf{Gap\,$\downarrow$} & \textbf{G-Ins} & \textbf{Arena} \\
\midrule

% --- Group 1: Baseline ---
\multicolumn{6}{l}{\textit{Baseline: Qwen3-4B-SFT}} \\
\hspace{1em} Zero-shot CoT           & 37.11 & 27.29 & 9.82  & 32.83 & 13.0 \\
\hspace{1em} Tool-use CoT      & 89.10 & 73.33 & 15.77 & 92.06 & 88.0 \\

\midrule

% --- Group 2: GRAIN (Ours) ---
\rowcolor{grainhighlight}
\multicolumn{6}{l}{\textbf{GRAIN-4B (Ours)}} \\
\rowcolor{grainhighlight}
\hspace{1em} \textbf{SFT + ARPO} & \textbf{95.76} & \textbf{87.96} & \textbf{7.80} & \textbf{97.05} & \textbf{91.0} \\

\bottomrule
\end{tabular*}
\caption{\textbf{OOD Robustness \& Generalization.} We report the accuracy drop ($\Delta$) on GRIT and zero-shot performance on unseen benchmarks (GraphInstruct, GraphArena).}
\label{tab:3}

\end{table}

\paragraph{Empowering Compact Models.} 
GRAIN also unlocks the reasoning potential of smaller architectures. While Llama-3.2-Instruct fails in zero-shot and standard tool-use settings ($<$5\%), GRAIN boosts its average accuracy to \textbf{47.41\%}, matching the performance of the proprietary GPT-5-nano (47.45\%). This indicates that our framework effectively instills structural reasoning capabilities even within compact parameter spaces.
    
\subsubsection{Robustness against Naming and Formulation Shifts}
\label{sec:exp_robustness}

Figure~\ref{fig:4} illustrates stability across eight variations. GRAIN demonstrates exceptional \textbf{structural invariance}, maintaining a nearly uniform performance envelope across all axes. Conversely, SFT-Tool exhibits marked \textbf{node-label sensitivity}, with performance visibly collapsing under ``Random'' and ``Mixed'' identifiers. This confirms GRAIN effectively decouples reasoning from surface forms, mitigating the overfitting to canonical patterns inherent in standard SFT. Further analysis of the two sensitivities are presented in the Appendix~\ref{sec:pilot_study}.

\definecolor{grainhighlight}{HTML}{F0F6F6}
\begin{table}[t]
\centering
\small
\renewcommand{\arraystretch}{1.25}
\setlength{\tabcolsep}{0pt} 

\begin{tabular*}{\columnwidth}{@{\extracolsep{\fill}} l c c cc c }
\toprule
\multicolumn{1}{c}{\multirow{2}{*}{\textbf{Method}}} & 
\textbf{Rounds} & 
\textbf{Tokens} & 
\multicolumn{2}{c}{\textbf{Latency (s)}} & 
\textbf{Acc.} \\
\cmidrule(lr){4-5}
 & (Avg) & (Avg) & \textbf{Avg} & \textbf{p90} & \textbf{(\%)} \\
\midrule

% --- Group 1: Single Agent ---
\multicolumn{6}{l}{\textit{GPT-5-nano (Single-Agent)}} \\
\hspace{1em} Zero-shot CoT & 1.00 & 17.1k & 117.3 & 200.2 & 29.95 \\
\hspace{1em} Tool-use CoT  & 2.00 & 22.1k & 126.1 & 212.8 & 47.45 \\

% --- Group 2: Multi-Agent ---
\multicolumn{6}{l}{\textit{Multi-Agent Framework}} \\
\hspace{1em} MA-GTS (4o-mini) & 6.00 & 13.5k & 58.4  & 96.1  & 79.31 \\

\midrule

% --- Group 3: Ours ---
\rowcolor{grainhighlight}
\multicolumn{6}{l}{\textbf{GRAIN-4B (Ours)}} \\
\rowcolor{grainhighlight}
\hspace{1em} \textbf{SFT + ARPO} & 
3.29$^\dagger$ & 
\textbf{3.3k} & 
\textbf{44.4} & 
\textbf{82.5} & 
\textbf{95.76} \\

\bottomrule
\multicolumn{6}{l}{\scriptsize \rule{0pt}{3ex}$^\dagger$ Denotes interaction rounds with the local Python environment.} \\
\end{tabular*}
\caption{GRAIN achieves peak accuracy (95.76\%) with the lowest overhead (3.3k tokens, 44.4s), significantly outperforming proprietary and multi-agent baselines.}
\label{tab:4}

\end{table}

\begin{figure}[t]
\vspace{-5pt}
\centering
  \includegraphics[width=0.5\textwidth]{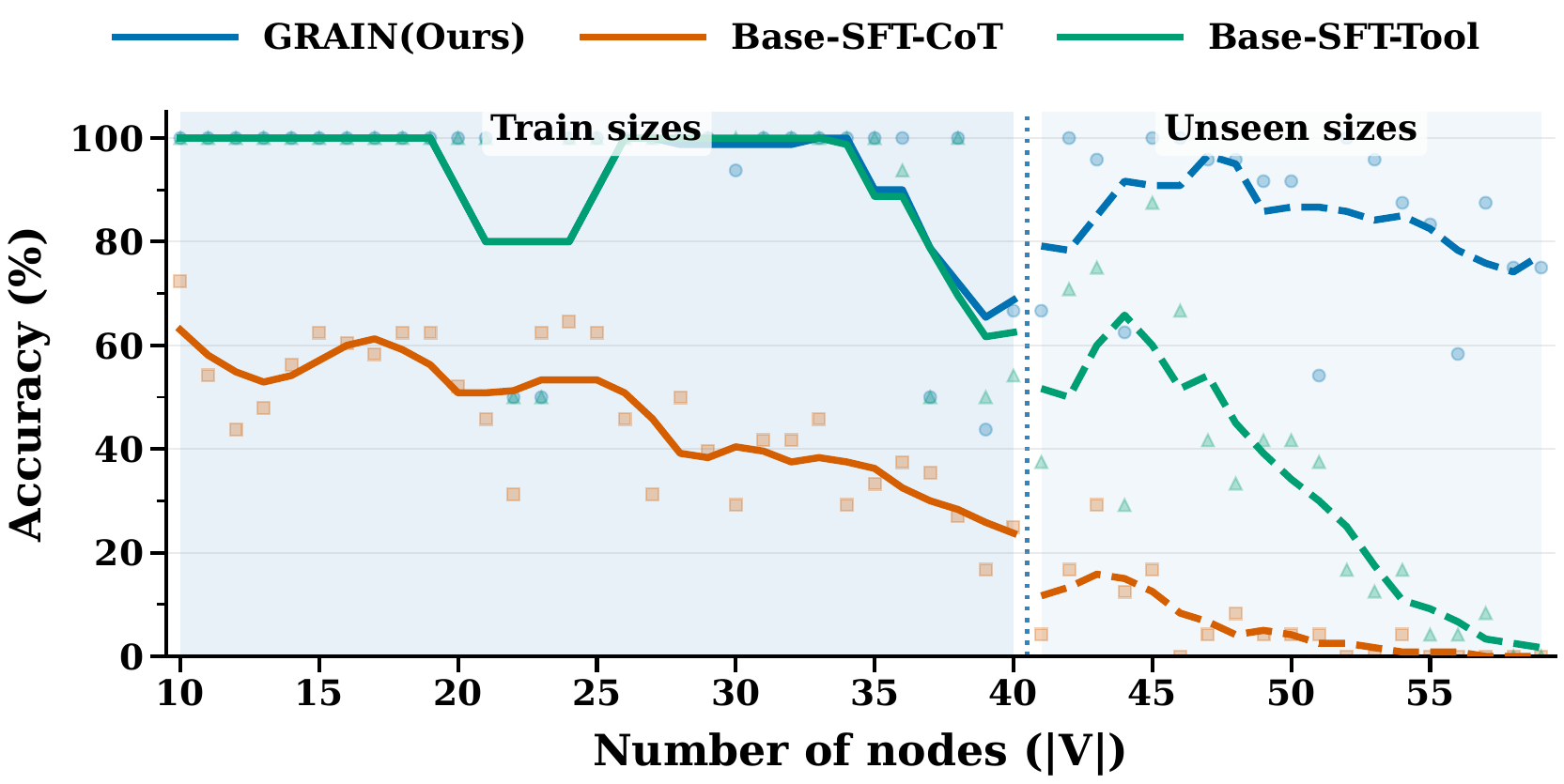}
 \caption{Performance as node count ($|V|$) exceeds the training range ($|V| \le 40$). GRAIN demonstrates robust length generalization, sustaining accuracy on larger graphs where baselines collapse.}
    \label{fig:5} 

\end{figure}

\definecolor{grainhighlight}{HTML}{F0F6F6}
\begin{table}[t]
\centering
\small
\renewcommand{\arraystretch}{1.3}
\setlength{\tabcolsep}{0pt}
\begin{tabular*}{\columnwidth}{@{\extracolsep{\fill}} l c cc c }
\toprule
\multirow{2}{*}{\textbf{Method / Variant}} & 
\multicolumn{3}{c}{\textbf{Metrics}} \\
\cmidrule(l){2-4}
 & \textbf{Gap\,$\downarrow$} & \textbf{Var. ($10^{-5}$)\,$\downarrow$} & \textbf{Acc.\,$\uparrow$} \\
\midrule

% --- GRAIN (Full) ---
% Gap: 1.14%, Var: 4.64e-5, Acc: 96.76%
\rowcolor{grainhighlight}
\textbf{GRAIN (Full)} & \textbf{1.14\%} & \textbf{4.64} & \textbf{96.76\%} \\

\midrule
\multicolumn{4}{l}{\textit{Component Analysis}} \\

% Row 3 in screenshot: w/o <graph> reward
% Gap: 11.14%, Var: 0.000146 -> 14.65, Acc: 84.73%
\hspace{1em} w/o Struct. Reward & 11.14\% & 14.65 & 84.73\% \\

% Row 2 in screenshot: w/o Full Name
% Gap: 4.11%, Var: 0.003418 -> 341.83, Acc: 79.21%
\hspace{1em} w/o Diverse Naming & 4.11\% & 341.83 & 79.21\% \\

\midrule
\multicolumn{4}{l}{\textit{Optimization Method}} \\

% Row 4 in screenshot: GRPO
% Gap: 1.17%, Var: 7.05e-5 -> 7.05, Acc: 96.51%
\hspace{1em} replace w/ GRPO & 1.17\% & 7.05 & 96.51\% \\

\bottomrule
\end{tabular*}
\caption{\textbf{Ablation Study.} \textbf{Gap} denotes the performance drop ($|Standard - Real|$), and \textbf{Var.} ($10^{-5}$) measures stability across naming permutations.}
\label{tab:ablation}
\end{table}

\subsubsection{OOD Generalization and Scalability}
\label{sec:exp_ood}

We evaluate GRAIN's capability to generalize beyond its training distribution across two dimensions: domain shifts (Table~\ref{tab:3}) and graph scale (Figure~\ref{fig:5}).

\paragraph{Distribution Shift and Transferability.} 
Table~\ref{tab:3} highlights GRAIN's resilience: while SFT-Tool drops $15.77\%$ on OOD splits due to overfitting, GRAIN halves this gap to \textbf{7.80\%}. Furthermore, GRAIN demonstrates exceptional \textbf{zero-shot transferability} on GraphInstruct (97.05\%) and GraphArena (91.0\%). To verify baseline fairness, GRAIN achieves \textbf{86.60\%} zero-shot accuracy on the unseen \textbf{G-REAL} dataset (MA-GTS's original benchmark; Appendix~\ref{sec:appendix_g_real}). This confirms that our invariance-oriented rewards foster universally applicable structural extraction skills.

\paragraph{Length Generalization.} 
Figure~\ref{fig:5} illustrates performance as graph size scales beyond the training range ($|V| > 40$). While baselines degrade rapidly on larger graphs due to context overload, GRAIN maintains robust accuracy. This \textbf{length generalization} is a direct benefit of our explicit pipeline design: by offloading computational complexity to external algorithms, GRAIN limits the LLM's role to semantic parsing and information extraction, effectively decoupling performance from the exponential complexity of graph reasoning.

\subsubsection{Ablation Study}
Table~\ref{tab:ablation} validates component contributions. Removing the \textbf{Structure Invariance Reward} widens the generalization gap ($1.14\% \to 11.14\%$) and drops accuracy to $84.73\%$, confirming its role in preventing overfitting. \textbf{Diverse Naming} is crucial for stability; omitting it spikes variance ($4.64 \to 341.83$) and degrades accuracy to $79.21\%$. Finally, \textbf{ARPO} outperforms GRPO by lowering variance ($7.05 \to 4.64$) and boosting accuracy ($96.51\% \to 96.76\%$), demonstrating the efficacy of branch-aware optimization.

\begin{table}[t]
\centering
\small
\renewcommand{\arraystretch}{1.2}
\setlength{\tabcolsep}{0pt}
\begin{tabular*}{\columnwidth}{@{\extracolsep{\fill}} l c c c}
\toprule
\textbf{Method / Variant} & \textbf{Node-F1} & \textbf{Edge-F1} & \textbf{Exact} \\
\midrule
\hspace{1em} w/o Struct. Reward & 96.03 & 94.95 & 63.10 \\
\hspace{1em} SFT-Tool & 98.45 & 97.96 & 90.21 \\
\hspace{1em} replace w/ GRPO & 97.23 & 97.20 & 95.22 \\
\rowcolor{grainhighlight}
\textbf{GRAIN (Full)} & \textbf{98.76} & \textbf{98.67} & \textbf{97.28} \\
\bottomrule
\end{tabular*}
\caption{\textbf{Intermediate Structure Recovery.} Results on 2,760 GRIT examples. Edge-F1 denotes weighted edge recovery, while Exact requires the entire predicted weighted graph to match the ground-truth graph.}
\label{tab:structure_recovery}
\end{table}

\paragraph{Intermediate Structure Recovery.}
As shown in Table~\ref{tab:structure_recovery}, GRAIN achieves \textbf{98.67\%} weighted Edge-F1 and \textbf{97.28\%} exact graph recovery. Removing the Structure Invariance Reward reduces exact recovery to \textbf{63.10\%} despite retaining high node- and edge-level overlap, showing that partial overlap can mask whole-graph inconsistencies. Replacing ARPO with GRPO also reduces exact recovery from $97.28\%$ to $95.22\%$. Appendix~\ref{sec:appendix_structure_errors} provides a detailed decomposition of the remaining non-exact predictions.

\subsubsection{Efficiency Analysis}
\label{sec:exp_efficiency}
Table~\ref{tab:4} highlights GRAIN's optimal accuracy-efficiency trade-off.

\paragraph{Deployment Efficiency.} 
Compared to GPT-5-nano's verbose reasoning ($>$17k tokens), GRAIN's concise structured generation cuts token usage by $\sim$85\% (to \textbf{3.3k}) and achieves the lowest latency (\textbf{44.4s}).

\paragraph{Advantage over Multi-Agent Systems.} 
While MA-GTS requires 6 interaction rounds, GRAIN's single-agent design outperforms it by \textbf{16.45\%} in accuracy while reducing latency by \textbf{24\%}. Our failure analysis (Appendix~\ref{sec:appendix_failure_analysis}) reveals that multi-agent systems frequently suffer from fatal information loss during handovers. Thus, for structural semantic parsing, GRAIN delivers SOTA performance without the prohibitive communication penalties of multi-agent architectures.

\section{Conclusion}

We address LLM fragility in graph reasoning under identifier and task shifts. We propose \textbf{GRAIN}, a single-agent RL framework utilizing a \textbf{Structure Invariance Reward} to decouple topology from surface variations. On our \textbf{GRIT} benchmark, GRAIN achieves state-of-the-art accuracy and efficiency, outperforming complex multi-agent systems. Furthermore, it generalizes robustly to OOD scenarios and larger graphs, validating the efficacy of invariant structural grounding.

\section*{Limitations}

A limitation of our current framework lies in the management of tool contexts. Given the vast array of graph algorithms, embedding comprehensive tool definitions directly into the system prompt leads to excessive text length, which consumes valuable context window and may distract the model from core reasoning tasks. Future work will explore adopting standards like the Model Context Protocol (MCP)~\citep{hou2025model} to handle tool definitions dynamically, moving away from static inclusion to optimize context efficiency.

\section*{Acknowledgments}

The research in this article is supported by the National Science Foundation of China (U22B2059, 62276083), Key Research and Development Program of Heilongjiang Province (2024ZX01A05) and the 5G Application Innovation Joint Research Institute's Project (A003).

\bibliography{custom}
\clearpage
\appendix
\raggedbottom

% Compact appendix float layout: keep wide tables with nearby discussion and
% avoid short float-only pages while preserving the main-text spacing.
\setcounter{topnumber}{4}
\setcounter{dbltopnumber}{4}
\renewcommand{\topfraction}{0.98}
\renewcommand{\dbltopfraction}{0.98}
\renewcommand{\textfraction}{0.02}
\renewcommand{\floatpagefraction}{0.85}
\renewcommand{\dblfloatpagefraction}{0.85}
\setlength{\textfloatsep}{6pt plus 2pt minus 2pt}
\setlength{\dbltextfloatsep}{6pt plus 2pt minus 2pt}
\setlength{\floatsep}{6pt plus 2pt minus 2pt}
\setlength{\dblfloatsep}{6pt plus 2pt minus 2pt}
\captionsetup[table]{skip=3pt}

\section{GRIT Benchmark Implementation Details}
\label{sec:app_grit}

This appendix complements the statistics in Table~\ref{tab:1} and task descriptions in Table~\ref{tab:task_descriptions_appendix} by detailing the generation pipeline.

\paragraph{Graph Topology Generation.}
Underlying graphs are synthesized using two random models via NetworkX: \textbf{Erdős-Rényi} (edge probability $p \in [0.1, 0.3]$) . For weighted tasks, weights are integers uniformly sampled from $[1, 10]$. We perform rejection sampling to ensure connectivity (for undirected graphs) or strong connectivity (for TSP).

\paragraph{Ground Truth \& Tie-Breaking.}
To ensure deterministic evaluation, we apply specific constraints to algorithmic solvers:
\begin{itemize}
    \item \textbf{BFS Traversal:} Neighbors are visited in \textit{lexicographical order} (e.g., node 2 before node 10) to guarantee a unique traversal sequence.
    \item \textbf{TSP:} Formulated as Metric TSP on complete graphs; we compute the exact minimum-cost Hamiltonian cycle using dynamic programming solvers.
    \item \textbf{Graph Coloring:} We compute the exact chromatic number using a greedy strategy with backtracking.
\end{itemize}

\paragraph{Narrative Injection Pipeline.}
We implement a template-based slot-filling engine. For each of the 31 scenarios (Table~\ref{tab:task_descriptions_appendix}), we define a domain context (e.g., ``Logistics'') and specific entity lists. The engine injects graph topology into these templates under four naming schemes. For \textbf{Random IDs}, we use high-entropy integers (e.g., 8392, 1024) to test tokenization robustness; for \textbf{Noisy/Mixed}, we introduce aliases (referring to the same node by multiple names).

\paragraph{OOD Split Construction.}
Unlike random splits, the \textbf{OOD Test} set is constructed by explicitly \textbf{holding out} 24 specific scenario templates and naming distributions during training. For instance, if ``City Traffic'' appears in Train, ``Server Routing'' is reserved for OOD. This ensures that high performance reflects true structural reasoning transfer rather than domain text memorization.

Please refer to Table~\ref{tab:naming_schemes_examples} for illustrative examples of the node naming variations (Canonical, Random IDs, Semantic, and Noisy/Mixed) used in our robustness evaluation. The complete system prompt, detailing the tool-use schema and reasoning guidelines for our GRAIN framework, is listed in Table~\ref{tab:system_prompt}.

\section{Training Implementation Details}
\label{sec:app_training}

This appendix details the hyperparameters and infrastructure used for the two-stage training pipeline of GRAIN: (1) Supervised Fine-Tuning (SFT) and (2) Agentic Reinforcement Learning via ARPO.

\subsection{Infrastructure}
All experiments were conducted on a computational node equipped with \textbf{4$\times$ NVIDIA A100 (80GB) GPUs}. 
For memory-efficient full-parameter fine-tuning, we utilized \texttt{DeepSpeed} with ZeRO-3 Offload strategy. 
For the RL stage, we integrated the \texttt{vLLM} inference engine to enable high-throughput trajectory rollouts. 
The training was performed in \textbf{BF16} precision to maintain numerical stability.

\subsection{Stage 1: Supervised Fine-Tuning (SFT)}
The SFT stage initializes the model's ability to use tools and generate valid JSON graph representations. We perform full-parameter fine-tuning on the \textbf{Qwen-3-Base (4B)} and \textbf{Llama-3.2-Base (3B)} models.

As listed in Table~\ref{tab:sft_details} (refer to main text or appendix table), we utilize a global batch size of 8 and a conservative learning rate of $7 \times 10^{-6}$ with a cosine decay scheduler. Notably, we set a context window of \textbf{15,000 tokens} to accommodate the verbose serialization of larger graphs (up to 40 nodes) and the corresponding CoT reasoning.

\begin{table}[h]
\centering
\small
% 调整行间距，使表格看起来舒展不拥挤
\renewcommand{\arraystretch}{1.2}
% 设为0，依靠 extracolsep 自动填充单栏宽度
\setlength{\tabcolsep}{0pt}

\begin{tabular*}{\columnwidth}{@{\extracolsep{\fill}} l r }
\toprule
\textbf{Parameter} & \textbf{Value} \\
\midrule

% --- General Settings ---
\multicolumn{2}{l}{\textit{\textbf{General Settings}}} \\
\hspace{3mm} Base Model & Qwen-3-4B \\
\hspace{3mm} Tuning Method & Full Fine-tuning \\
\hspace{3mm} Precision & BF16 \\
\hspace{3mm} Context Window & 15,000 tokens \\
\hspace{3mm} DeepSpeed Stage & ZeRO-3 (Offload) \\
\hspace{3mm} GPUs & $2 \times$ NVIDIA A100 GPUs \\

\midrule

% --- Hyperparameters ---
\multicolumn{2}{l}{\textit{\textbf{Hyperparameters}}} \\
\hspace{3mm} Global Batch Size & 8 \\
\hspace{3mm} Gradient Accumulation & 2 \\
\hspace{3mm} Epochs & 3 \\
\hspace{3mm} Learning Rate & $7 \times 10^{-6}$ \\
\hspace{3mm} LR Scheduler & Cosine \\
\hspace{3mm} Warmup Ratio & 0.1 \\

\bottomrule
\end{tabular*}
\caption{\textbf{Implementation Details.} Hyperparameters and settings used for full-parameter fine-tuning.}
\label{tab:sft_details}
\end{table}

\begin{table*}[htbp]
\centering
\small % 保持小号字体

% 调整列格式：
% c : 第一列居中对齐
% >{\raggedright\arraybackslash}p{5cm} : 第二列左对齐，宽度减小到 5cm
% >{\raggedright\arraybackslash}p{7cm} : 第三列左对齐，宽度减小到 7cm
% 这样总宽度大约在 14-15cm 左右，可以很好地适应 ACL 的跨栏宽度而不超标。
\begin{tabular}{c >{\raggedright\arraybackslash}p{5cm} >{\raggedright\arraybackslash}p{7cm}}
\toprule
\textbf{Graph Task} & \textbf{Canonical Graph-Theoretic Formulation} & \textbf{Representative Real-World Problem Formulations (Noun Phrases)} \\
\midrule
Centrality & Compute centrality scores of one or several nodes in a graph and identify structurally ``key'' nodes. & City traffic hubs; emergency response centers; social network key persons; information relay stations \\ \addlinespace
Shortest Path & In a weighted directed or undirected graph, find the minimum-cost path from a source node to a target node. & Data-center latency routing; logistics cost routes; robot energy-efficient paths; network routing paths \\ \addlinespace
TSP & In a weighted complete graph, find a minimum-cost tour that visits every node once and returns to the start (decision version NP-complete, optimization version NP-hard). & Parcel delivery tour; sales representative tour; food delivery tour; garbage collection tour; maintenance inspection tour; school bus route \\ \addlinespace
Min Graph Coloring & Color all nodes of a graph using as few colors as possible so that adjacent nodes have different colors. & Wi-Fi channel assignment; cellular frequency planning; radio channel allocation; event loudspeaker placement; exam timetabling; map region coloring; interference-aware equipment layout \\ \addlinespace
Graph Traversal (BFS) & Perform breadth-first search from a given source node and output the visit order or layer structure. & Post-disaster inspection sweep; tourist exploration route; UAV area scanning; building safety inspection \\ \addlinespace
Vertex Cover & Select a minimum set of nodes such that every edge has at least one endpoint in the set. & Server monitoring placement; road intersection cameras; railway security posts; social liaison selection; pipeline sensor placement; campus patrol posts \\
\bottomrule
\end{tabular}
\caption{Detailed descriptions of the six graph reasoning tasks included in the GRIT benchmark. This table complements the main results by providing the canonical graph-theoretic formulation and representative real-world problem scenarios for each task.}
\label{tab:task_descriptions_appendix}
\end{table*}
\subsection{Stage 2: Reinforcement Learning (ARPO)}
We employ our proposed \textbf{ARPO} (Agentic Reasoning Policy Optimization) algorithm, utilizing the GRPO estimator. The model is warm-started from the tool-use SFT checkpoint. The detailed hyperparameters are provided in Table~\ref{tab:rl_details}.

\paragraph{Curriculum Warm-up Strategy.}
As mentioned in the experimental setup, we adopt a two-phase curriculum to ensure training stability:
\begin{enumerate}
    \item \textbf{Phase 1 (Stabilization):} We first train on a subset of smaller graphs ($N \in [4, 14]$). This phase focuses on stabilizing the model's adherence to the structured output format and tool invocation syntax without the distraction of long-context complexity.
    \item \textbf{Phase 2 (Main Optimization):} We then scale to the full training distribution ($N \in [10, 40]$). To balance diversity, we sample two distinct underlying graphs per size per task during this phase.
\end{enumerate}

\begin{table}[h]
\centering
\small
\renewcommand{\arraystretch}{1.2}
\setlength{\tabcolsep}{0pt}

\begin{tabular*}{\columnwidth}{@{\extracolsep{\fill}} l r }
\toprule
\textbf{Parameter} & \textbf{Value} \\
\midrule

% --- Group 1: Optimization ---
\multicolumn{2}{l}{\textit{\textbf{Optimization \& Data}}} \\
\hspace{3mm} Actor LR & $1 \times 10^{-6}$ \\
\hspace{3mm} Train Batch Size & 16 \\
\hspace{3mm} Mini-batch Size & 4 \\
\hspace{3mm} Epochs & 2 \\
\hspace{3mm} KL Coefficient & 0.0 \\

\midrule

% --- Group 2: Generation & Environment ---
\multicolumn{2}{l}{\textit{\textbf{Generation \& Environment}}} \\
\hspace{3mm} Rollout Engine & vLLM \\
\hspace{3mm} Max Prompt Length & 24,000 tokens \\
\hspace{3mm} Max Response Length & 12,000 tokens \\
\hspace{3mm} Rollouts per Prompt ($G$) & 8 \\
\hspace{3mm} Reward Function & Graph Correctness \\
\hspace{3mm} Tool Integration & Sync with Tool \\

\midrule

% --- Group 3: ARPO/Search Specifics ---
\multicolumn{2}{l}{\textit{\textbf{Search \& Exploration}}} \\
\hspace{3mm} Beam Size & 2 \\
\hspace{3mm} Branch Probability & 0.5 \\
\hspace{3mm} Entropy Weight & 0.2 \\
\hspace{3mm} Initial Rollouts & 2 \\

\bottomrule
\end{tabular*}
\caption{\textbf{Hyperparameters for RL Training (ARPO).} We employ the GRPO estimator with vLLM-based rollouts. Note the extended context window (36k total) to handle graph reasoning trajectories.}
\label{tab:rl_details}
\end{table}

\paragraph{Extended Context and Rollouts.}
Graph reasoning trajectories involving tool interactions significantly increase sequence length. Consequently, we extend the maximum context window to \textbf{36,000 tokens} (24k for prompts + 12k for generation). We set the KL coefficient to $0.0$, allowing the policy to explore the solution space freely, constrained only by the group-relative reward signal. The training uses a batch size of 16 with 8 rollouts per prompt ($G=8$) to stabilize the baseline estimation.

\begin{table*}[htbp]
\centering
\small % 使用小号字体

% 使用 tabular* 并设置宽度为 \textwidth，自动填充列间距
\begin{tabular*}{\textwidth}{@{\extracolsep{\fill}}llccccc}
\toprule
% 表头第一行：宏观分类
\multirow{2}{*}{\textbf{Task}} & \multirow{2}{*}{\textbf{Model}} & \multicolumn{4}{c}{\textbf{Node Naming Schemes (Accuracy \%)}} & \multirow{2}{*}{\textbf{Var ($\times 10^{-4}$)}} \\
% 表头第二行：具体命名方式
\cmidrule(lr){3-6}
& & \textbf{Canonical} & \textbf{Random ID} & \textbf{Semantic} & \textbf{Mixed} & \\

% 闭源模型部分
\midrule
\multicolumn{7}{c}{\textbf{\textit{Closed Source Models}}} \\
\midrule
\multirow{5}{*}{Synthetic Graphs}
& gpt-5-nano            & 83.33 & 66.67 & 75.00 & 70.00 & 52.50 \\
& deepseek-3.2-reason              & 96.61 & 91.53 & 93.22 & 96.67 & 6.54 \\
& Qwen-plus             & 80.39 & 82.00 & 82.35 & 84.31 & 2.59 \\
& gemini-2.5-flash-lite & 23.33 & 13.33 & 51.67 & 40.00 & 292.00 \\
& kimi-k2-thinking      & 37.21 & 23.26 & 41.86 & 22.73 & 94.80 \\
\cmidrule{1-7}
\multirow{5}{*}{Real-world Queries}
& gpt-5-nano            & 71.67 & 76.67 & 68.33 & 65.00 & 24.80 \\
& deepseek-3.2-reason              & 86.44 & 94.92 & 91.53 & 90.00 & 12.40 \\
& Qwen-plus             & 72.55 & 76.47 & 84.31 & 74.51 & 26.60 \\
& gemini-2.5-flash-lite & 56.67 & 68.33 & 65.00 & 56.67 & 35.10 \\
& kimi-k2-thinking      & 20.93 & 13.95 & 25.58 & 18.18 & 23.80 \\

% 开源模型部分
\midrule
\multicolumn{7}{c}{\textbf{\textit{Open Source Models}}} \\
\midrule
\multirow{3}{*}{Synthetic Graphs}
& Qwen3\_4b\_ins               & 61.67 & 51.67 & 58.33 & 60.00 & 19.20 \\
& Qwen3\_4b                    & 60.00 & 40.00 & 51.67 & 51.67 & 67.60 \\
& DeepSeek-R1-Distill-Qwen-7B  & 10.00 & 3.33  & 8.33  & 13.33 & 17.40 \\
\cmidrule{1-7}
\multirow{3}{*}{Real-world Queries}
& Qwen3\_4b\_ins               & 50.00 & 45.00 & 50.00 & 48.33 & 5.56 \\
& Qwen3\_4b                    & 53.33 & 43.33 & 46.67 & 48.33 & 17.40 \\
& DeepSeek-R1-Distill-Qwen-7B  & 10.00 & 8.33  & 3.33  & 8.33  & 8.34 \\
\bottomrule
\end{tabular*}
\caption{Performance comparison across different node naming schemes on Synthetic Graphs and Real-world Queries. Accuracy is reported in percentages, and variance (Var) is scaled by $10^{-4}$.} 
\label{tab:model_performance}
\end{table*}
\section{Qualitative Case Study}
\label{sec:app_case_study}

To provide a granular understanding of how \textbf{GRAIN} differs from existing paradigms, we present a qualitative comparison on a representative Shortest Path instance involving a real-world narrative with semantic noise (e.g., irrelevant details about ``taxis being unavailable''). We analyze the reasoning trajectories of GRAIN (Table~\ref{tab:case_study_fixed}), a proprietary model using CoT (GPT-5-nano, Table~\ref{tab:baseline_gpt5}), and a multi-agent framework (MA-GTS, Table~\ref{tab:baseline_magts}).

\paragraph{GRAIN: Decoupling Parsing from Computation.}
As shown in Table~\ref{tab:case_study_fixed}, GRAIN adopts a ``parse-then-solve'' approach. The model does not attempt to calculate the path distance internally. Instead, it focuses entirely on \textbf{semantic parsing}: filtering out the narrative noise and extracting the topological structure into a rigorous intermediate representation (the JSON adjacency list). Crucially, GRAIN recognizes that pathfinding is a computational task, not a linguistic one. By explicitly invoking the Dijkstra tool with the correct parameters, it guarantees arithmetic correctness. This case demonstrates GRAIN's core advantage: it uses the LLM solely for its strength (unstructured text understanding) while offloading algorithmic complexity to the external environment.

\paragraph{GPT-5-nano (CoT): The Fragility of Internal Simulation.}
Table~\ref{tab:baseline_gpt5} illustrates the reliance of standard LLMs on \textbf{internal simulation}. The model attempts to mimic the state transitions of Dijkstra's algorithm step-by-step within its context window (e.g., calculating ``Relax TB: $0+1=1$''). While successful in this small-scale example, this approach is inherently fragile. The model is forced to act as both a parser and an arithmetic engine. In real-world scenarios with larger graphs or floating-point weights, such ``mental simulation'' is prone to state tracking errors and calculation hallucinations, as the model lacks an external verifier for its intermediate arithmetic steps.

\paragraph{MA-GTS: Redundant Procedural Fragmentation.}
The Multi-Agent approach (Table~\ref{tab:baseline_magts}) correctly identifies the structure but suffers from \textbf{procedural fragmentation}. The task is decomposed into granular roles---an \textit{Info Extractor} finds entities, a \textit{Graph Builder} creates edges, and a \textit{Solver} executes code. While this reduces the cognitive load on any single agent, the case study reveals that for fundamental graph problems, such fragmentation is often unnecessary. The strict role boundaries require the explicit serialization and re-parsing of information between agents (e.g., passing entity lists from Agent 1 to Agent 2), introducing potential information loss at each handover. GRAIN demonstrates that a single, well-optimized agentic policy can internalize this pipeline, achieving the same structural accuracy without the fragmented decision process.

\paragraph{Summary of Paradigms.}
The comparison highlights a fundamental shift in the reasoning approach:
\begin{itemize}
    \item \textbf{CoT (GPT-5)} treats graph reasoning as a \textit{sequence prediction} problem, vulnerable to calculation errors.
    \item \textbf{Multi-Agent (MA-GTS)} treats it as an \textit{organizational} problem, leading to rigid and fragmented workflows.
    \item \textbf{GRAIN} treats it as a \textit{translation-and-execution} problem. By aligning natural language with symbolic intermediate representations, GRAIN achieves the robustness of tool-augmented systems while maintaining a streamlined, single-agent decision boundary.
\end{itemize}

\section{Empirical Study: Node-Naming Robustness on Synthetic and Real-World Graphs}
\label{sec:pilot_study}

Before detailing our method, we quantify the node-label sensitivity of LLMs using the Shortest Path task across varying naming schemes (Canonical, Random ID, Semantic, Mixed). As shown in Figure~\ref{fig:2} and Table~\ref{tab:model_performance}, three trends validate the need for robust grounding:

\paragraph{High Volatility.} The substantial error bars across all models highlight severe instability. Mere surface-level changes in node identifiers or narrative framings induce drastic performance fluctuations, indicating that current LLMs lack consistent structural understanding.

\paragraph{Pattern Dependency in Open Models.} Open-source models (Fig.~\ref{fig:2}, bottom) suffer a sharp performance drop on \textit{Random IDs} compared to Semantic or Canonical names. This suggests a reliance on sequential text patterns or meaningful words rather than the underlying topology.

\paragraph{Semantic Overfitting in Closed Models.} Surprisingly, on real-world graphs (Fig.~\ref{fig:2}, top-right), closed-source models perform slightly worse on \textit{Canonical} names ($v_1 \dots v_n$) than on \textit{Semantic} names. This implies an over-fitting to rich contexts, where stripping a problem down to abstract symbols paradoxically hinders performance.

Qualitative analysis confirms that these failures predominantly occur during the \textbf{grounding phase} (mapping text to graph representations), motivating GRAIN's focus on invariant decision-making.

\begin{table*}[t]
    \centering
\
    \resizebox{\textwidth}{!}{%
    \begin{tabular}{l p{5.5cm} p{8.5cm}}
        \toprule
        \textbf{Naming Scheme} & \textbf{Definition \& Characteristics} & \textbf{Representative Text Realization (Example)} \\
        \midrule
        \textbf{Canonical} & 
        Uses standard graph-theoretic indices (e.g., integers $0, 1$) or generic labels ($v_i$). Represents the sanitized format typical of academic benchmarks. & 
        \textit{"There is an edge connecting \textbf{Node 0} to \textbf{Node 1} with a weight of 7. Find the path from 0 to 1."} \\
        \midrule
        \textbf{Random IDs} & 
        Assigns high-entropy, non-sequential integers (e.g., from $[1, 10000]$). Tests robustness against tokenization fragmentation and lack of numerical continuity. & 
        \textit{"Server \textbf{8392} is connected to Server \textbf{104} with latency 7ms. Start routing from \textbf{8392}."} \\
        \midrule
        \textbf{Semantic} & 
        Uses meaningful, domain-specific entity names (e.g., locations, proteins). Introduces semantic priors that may distract the model from topological reasoning. & 
        \textit{"Walk from \textbf{Xundral View} to \textbf{Knights Market}. The distance is 7 km. Identify the route departing from \textbf{Xundral View}."} \\
        \midrule
        \textbf{Noisy / Mixed} & 
        Simulates "dirty" real-world data by mixing aliases, \textbf{meaningless strings (e.g., hashes, garbled text)}, and heterogeneous types. Tests entity resolution and resilience to surface-form noise. & 
        \textit{"Link: \textbf{'0x9A\_err'} $\rightarrow$ \textbf{unknown\_loc} (dist: 7). Note: \textbf{'0x9A\_err'} is also referred to as \textbf{Start\_Pt}. Avoid nodes marked as \textbf{\#\#@\$}."} \\
        \bottomrule
    \end{tabular}%
    
    }
        \caption{\textbf{Examples of Node Naming Schemes in GRIT.} We illustrate how the same underlying edge (from a source node to a target node with weight 7) is realized textually across four different naming schemes. This highlights the spectrum from clean, sanitized inputs to noisy, unstructured real-world data.}
    \label{tab:naming_schemes_examples}
\end{table*}

% =========================================================================================
% 依赖包 (如果主文件中没有引入，请取消注释)
% \usepackage{tabularx}
% \usepackage{booktabs}
% \usepackage{listings}
% \usepackage{xcolor}
% =========================================================================================

% --- 颜色定义 (保持与你之前的表格一致) ---
\definecolor{mygreen}{RGB}{0, 100, 0}
\definecolor{myblue}{RGB}{0, 0, 139}
\definecolor{mypurple}{RGB}{128, 0, 128}
\definecolor{jsonkey}{RGB}{30, 50, 160}
\definecolor{jsonval}{RGB}{160, 30, 30}
\definecolor{cmdgray}{RGB}{100, 100, 100}

% --- JSON 样式定义 ---
\lstdefinelanguage{jsonPrompt}{
    basicstyle=\normalfont\ttfamily\footnotesize,
    string=[s]{"}{"},
    stringstyle=\color{jsonval},
    comment=[l]{:},
    commentstyle=\color{black},
    keywordstyle=\color{jsonkey}\bfseries,
    breaklines=true,
    breakatwhitespace=true,
    showstringspaces=false,
    upquote=true,
    framesep=0pt,
    rulesep=0pt,
    frame=none
}

% --- 定义保存盒子 (防止与之前的 jsonbox 冲突，这里命名为 promptjsonbox) ---
\newsavebox{\promptjsonbox}

\begin{table*}[h]
\centering
\footnotesize
\renewcommand{\arraystretch}{1.15}

% --- 【关键步骤】预渲染 JSON Template ---
\begin{lrbox}{\promptjsonbox}
\begin{minipage}{0.85\textwidth}
\begin{lstlisting}[language=jsonPrompt]
{
  "problem_type": "problem_type_here",
  "graph": {
    "type": "adjacency_list",
    "data": {
      "node1": {
        "neighbor1": weight1,
        "neighbor2": weight2
      },
      "node2": {
        "neighbor3": weight3,
        "neighbor4": weight4
      }
    },
    "directed": true_or_false,
    "weighted": true_or_false
  },
  "parameters": {
    "parameter1": "parameter1_value",
    "parameter2": "parameter2_value",
    "algorithm": "algorithm_name"
  }
}
\end{lstlisting}
\end{minipage}
\end{lrbox}
% --- 盒子定义结束 ---

\begin{tabularx}{\textwidth}{lX}
\toprule
\textbf{Component} & \textbf{Content / Instruction} \\
\midrule

% --- Role Definition ---
\textbf{Role Definition} & 
You are a helpful assistant that specializes in solving graph theory problems using a \textbf{graph theory tool}. Given a question, you need to first think about the reasoning process in your mind and then provide the answer. During thinking, you can invoke the graph theory tool to compute shortest paths, minimum spanning trees, topological sorting, or other graph algorithms if needed. \\
\midrule

% --- Interaction Protocol ---
\textbf{Protocol \& Tags} & 
The reasoning process and answer must be enclosed within specific XML tags. Follow these guidelines strictly:
\begin{enumerate}
    \item \textbf{Thinking:} Start with \textcolor{mygreen}{\texttt{<think>}} and explain your step-by-step reasoning process. End with \textcolor{mygreen}{\texttt{</think>}}.
    \item \textbf{Tool Call:} When you need to use the graph theory tool, output \textcolor{myblue}{\texttt{<graph>}} followed by a JSON-formatted query, then \textcolor{myblue}{\texttt{</graph>}}.
    \item \textbf{Tool Result:} After the tool call, output \textcolor{mypurple}{\texttt{<result>}} followed by the tool's output, then \textcolor{mypurple}{\texttt{</result>}}.
    \item \textbf{Final Answer:} Finally, output \texttt{<answer>} followed by the final answer, then the final answer is \texttt{\textbackslash boxed\{answer here\}}\texttt{</answer>}.
\end{enumerate} \\
\midrule

% --- JSON Schema ---
\textbf{Tool Schema} & 
The JSON query inside \textcolor{myblue}{\texttt{<graph>}} tags must follow this exact structure:
\par\vspace{5pt}
\usebox{\promptjsonbox} % 插入预渲染的 JSON
\par\vspace{5pt}
Ensure that \texttt{problem\_type}, \texttt{graph structure}, and \texttt{algorithm} parameters are correctly populated based on the user query. \\
\bottomrule
\end{tabularx}

\caption{\textbf{System Prompt for GRAIN (SFT \& RL).} This prompt instructs the model to act as a graph reasoning agent, defining the explicit "Think-Tool-Result" loop and the strict JSON schema required for interacting with the deterministic graph solver.}
\label{tab:system_prompt}
\vspace{0.35em}

% 颜色定义

% Share one float page with Table~\ref{tab:system_prompt}.
\scriptsize
\renewcommand{\arraystretch}{1.05}
\setlength{\tabcolsep}{4pt}

\caption{\textbf{Summary of Notations.} Key symbols used in GRAIN. We update the reward notations to reflect the gated formulation and structure similarity score.}
\label{tab:notations_unified}

\begin{tabularx}{\textwidth}{@{} l X c l X @{}}
\toprule
\textbf{Symbol} & \textbf{Description} & \phantom{sp} & \textbf{Symbol} & \textbf{Description} \\
\midrule

% --- Part 1: Problem Formulation (左) vs Rewards (右) ---
\multicolumn{2}{l}{\textit{\textbf{Problem Formulation \& Environment}}} & & \multicolumn{2}{l}{\textit{\textbf{Rewards \& Optimization}}} \\

$G = (V, E, w)$          & Ground-truth weighted graph. & & $r_{\text{ans}}(\tau)$   & Answer correctness reward. \\
$\mathcal{T}$            & Graph reasoning task (e.g., TSP). & & $s_{\text{inv}}(\tau)$   & Structure similarity score ($\hat{G}$ vs. $G$). \\
$\nu: V \rightarrow \Sigma^{*}$ & Node naming function. & & $R(\tau)$                & Gated total return (Eq.~\ref{eq:gated_reward}). \\
$\phi$                   & Query formulation style. & & $\rho_{\text{err}}$      & Penalty constant for format violation. \\
$x$                      & Input query $x = f(G, \nu, \phi, \mathcal{T})$. & & $\lambda_{\text{inv}}$   & Coef.\ for structure similarity score. \\
$\mathcal{G}$            & Distribution of training graphs. & & $\beta$                  & Coef.\ for KL divergence penalty. \\
$\mathcal{P}(G)$         & Naming/style distribution. & & $J(\theta)$              & Total RLVR objective function. \\
$\hat{G}$                & Reconstructed candidate graph. & & & \\
$\hat{z}$                & Tool execution result. & & & \\

\addlinespace[0.5em] 

% --- Part 2: Agent & Trajectory (左) ---
\multicolumn{2}{l}{\textit{\textbf{Agent \& Trajectory}}} & & \multicolumn{2}{l}{} \\ 

$\pi_{\theta}$           & Policy (LLM) parameterized by $\theta$. & & & \\
$\pi_{\text{ref}}$       & Reference policy (SFT model). & & & \\
$\tau$                   & Generated trajectory sequence. & & & \\
$y$                      & Structured output with tags. & & & \\

\bottomrule
\end{tabularx}
\end{table*}

% 颜色定义
\definecolor{mygreen}{RGB}{0, 100, 0}
\definecolor{myblue}{RGB}{0, 0, 139}
\definecolor{mypurple}{RGB}{128, 0, 128}
\definecolor{jsonkey}{RGB}{30, 50, 160}
\definecolor{jsonval}{RGB}{160, 30, 30}

% JSON 样式定义
\lstdefinelanguage{json}{
    basicstyle=\normalfont\ttfamily\footnotesize,
    string=[s]{"}{"},
    stringstyle=\color{jsonval},
    comment=[l]{:},
    commentstyle=\color{black},
    keywordstyle=\color{jsonkey}\bfseries,
    breaklines=true,
    breakatwhitespace=true,
    showstringspaces=false,
    upquote=true,
    framesep=0pt,
    rulesep=0pt,
    frame=none
}

% 【关键】定义一个保存盒子，用于存放 JSON 代码
\newsavebox{\jsonbox}

\begin{table*}[t]
\centering
\footnotesize
\renewcommand{\arraystretch}{1.1}

% --- 【关键步骤】 ---
% 在进入 tabularx 之前，先将代码块渲染并存入 \jsonbox 盒子中
% minipage 宽度设为 0.82\textwidth，给左边的一列留出空间，防止右边溢出
\begin{lrbox}{\jsonbox}
\begin{minipage}{0.82\textwidth}
\begin{lstlisting}[language=json]
{
  "problem_type": "shortest_path",
  "graph": {
    "type": "adjacency_list",
    "data": {
      "Knights Market": {
        "Xundral View": 7,
        "Skyline Gardens": 1,
        "Dragons Gate": 3,
        "Trelvon Bay": 6
      },
      "Xundral View": { "Trelvon Bay": 1 },
      "Draycon Arcade": {
        "Copper Clock Square": 2,
        "Vorlen Heights": 8
      },
      "Copper Clock Square": {
        "Trelvon Bay": 7,
        "Vorlen Heights": 5,
        "Zaphrel Summit": 8
      },
      "Granite Plaza": { "Trelvon Bay": 8 },
      "Dragons Gate": { "Trelvon Bay": 7 },
      "Trelvon Bay": {}, 
      "Zaphrel Summit": {}
    },
    "directed": false, 
    "weighted": true
  },
  "parameters": {
    "start": "Xundral View",
    "end": "Zaphrel Summit",
    "algorithm": "dijkstra"
  }
}
\end{lstlisting}
\end{minipage}
\end{lrbox}
% --- 【结束盒子定义】 ---

\begin{tabularx}{\textwidth}{lX}
\toprule
\textbf{Stage} & \textbf{Content} \\
\midrule

% --- User Query ---
\textbf{User Query} & 
My younger sibling isn't feeling well, and the nearest hospital is at \textbf{Zaphrel Summit}. Right now we're at home, near an intersection I'll call \textbf{Xundral View}. [...] I checked a city map and noted down the walking distances: Knights Market to Xundral View is 7 km... \textit{[narrative omitted for brevity]} ... Dragons Gate to Trelvon Bay is 7 km. \textbf{Starting from Xundral View, what route should we walk to reach Zaphrel Summit while minimizing the total walking distance?} \\ 
\midrule

% --- Reasoning ---
\textbf{Reasoning} & 
\textcolor{mygreen}{\texttt{<think>}} From an application perspective, this requires graph theory tool calculation. \textbf{Graph parameters identified:}
\begin{itemize}
    \item \textbf{Nodes:} 9 distinct locations (e.g., Knights Market, Xundral View).
    \item \textbf{Goal:} Compute shortest path from \textit{Xundral View} to \textit{Zaphrel Summit}.
    \item \textbf{Edge Extraction:} Parsing narrative to build adjacency list.
\end{itemize}
Tool call explanation: I will input graph information in JSON format, including nodes, edges, start and end points. \textcolor{mygreen}{\texttt{</think>}} \\ 
\midrule

% --- Action ---
\textbf{Action} & 
\textcolor{myblue}{\texttt{<graph>}}
\par\vspace{3pt}
% 直接使用刚才定义好的盒子，这里就像贴一张图一样安全，绝对不会乱
\usebox{\jsonbox}
\par\vspace{3pt}
\textcolor{myblue}{\texttt{</graph>}} \\ 
\midrule

% --- Observation ---
\textbf{Observation} & 
\textcolor{mypurple}{\texttt{<result>}}
Shortest path algorithm (Dijkstra) result:
\begin{itemize}
    \item \textbf{Path:} Xundral View $\to$ Trelvon Bay $\to$ Copper Clock Square $\to$ Zaphrel Summit
    \item \textbf{Total Length:} 16
\end{itemize}
\textcolor{mypurple}{\texttt{</result>}} \\ 
\midrule

% --- Final Response ---
\textbf{Final Response} & 
\textcolor{mygreen}{\texttt{<think>}} Tool call successful. Calculation result: shortest path contains 3 nodes... total weight 16. \textcolor{mygreen}{\texttt{</think>}} \newline
\textbf{Answer:} \fbox{16} \\ 
\bottomrule
\end{tabularx}
\caption{\textbf{Case Study on Shortest Path Task.} The model parses a noisy narrative, constructs a structured JSON representation, and invokes the tool to find the optimal path. The code block demonstrates the explicit structural constraints imposed by GRAIN.}
\label{tab:case_study_fixed}
\end{table*}

\begin{table*}[t]
\centering
\footnotesize
\renewcommand{\arraystretch}{1.08}
\begin{tabularx}{\textwidth}{lX}
\toprule
\textbf{Stage} & \textbf{Content} \\
\midrule

% --- User Query (Same as before) ---
\textbf{User Query} & 
My younger sibling isn't feeling well, and the nearest hospital is at \textbf{Zaphrel Summit}. Right now we're at home, near an intersection I'll call \textbf{Xundral View}. [...] \textit{(Narrative containing distances: Knights Market to Xundral View is 7 km, etc.)} ... \textbf{Starting from Xundral View, what route should we walk to reach Zaphrel Summit while minimizing the total walking distance?} \\ 
\midrule

% --- Model Response (CoT) ---
\textbf{Model Response} & 
\textcolor{mygreen}{\texttt{<think>}} 
\textbf{Task restatement:} Starting from Xundral View (XV), find a walking route to Zaphrel Summit (ZS) minimizing total distance... The graph is undirected and unweighted in terms of edges with weights equal to distances.

\vspace{4pt}
\textbf{1. Internal Graph Construction:}
\begin{itemize}
    \item Vertex set V = \{XV, KM, SG, DA, GP, DG, TB, CCS, VH, ZS\}.
    \item \textbf{Edge set extraction:}
    \begin{itemize}
        \item (XV, TB) = 1, (KM, XV) = 7, (KM, SG) = 1, (KM, DA) = 2
        \item (TB, CCS) = 7, (TB, GP) = 8, (CCS, ZS) = 8 ... \textit{[full list extracted]}
    \end{itemize}
\end{itemize}

\vspace{4pt}
\textbf{2. Algorithm Simulation (Dijkstra):}
Method: Run Dijkstra’s algorithm from XV, tracking tentative distances.
\begin{itemize}
    \item \textit{Initialize:} dist(XV) = 0; all others = $\infty$.
    \item \textit{Step 1:} From XV, relax TB with dist $0 + 1 = 1$. $\to$ \textbf{dist(TB) = 1}.
    \item \textit{Step 2 (Next smallest TB):}
    \begin{itemize}
        \item Relax KM: $1 + 6 = 7 \to$ dist(KM) = 7
        \item Relax CCS: $1 + 7 = 8 \to$ dist(CCS) = 8
        \item ... \textit{[relaxing other neighbors]}
    \end{itemize}
    \item \textit{Step 3 (Next smallest KM):} Relax SG via KM: $7 + 1 = 8 \to$ dist(SG) = 8...
    \item \textit{[...omitting intermediate relaxation steps for brevity...]}
    \item \textit{Final Step:} ZS reached with dist(ZS) = 16. No other path reduces this.
\end{itemize}

\textbf{Conclusion:} The minimal walking distance is 16.
\textcolor{mygreen}{\texttt{</think>}}
\par\vspace{5pt}
\textbf{Answer:} \fbox{16} \\ 
\bottomrule
\end{tabularx}
\caption{\textbf{Baseline Case Study: GPT-5-nano (Chain-of-Thought).} Unlike GRAIN which offloads computation to external tools, the baseline model attempts to \textbf{internally simulate} the execution of Dijkstra's algorithm via text generation. While successful in this instance, this approach relies heavily on the model's ability to maintain long-context state and perform precise arithmetic without external verification, which typically degrades on more complex graphs.}
\label{tab:baseline_gpt5}
\vspace{0.35em}

% Share one float page with Table~\ref{tab:baseline_gpt5}.
\scriptsize
\renewcommand{\arraystretch}{1.08}
\begin{tabularx}{\textwidth}{lX r} % 增加了一个右对齐的 Cost 列
\toprule
\textbf{Agent Role} & \textbf{Action \& Output Summary} & \textbf{Cost (Time/Tok)} \\
\midrule

% --- User Query ---
\textbf{User Query} & 
\multicolumn{2}{p{0.8\textwidth}}{
I'm a university student living near \textbf{Lighthouse Point}... final exam at \textbf{Umbra Theater}... Yondel Crossing to Lighthouse Point is 8 km... \textit{[...narrative omitted...]} ... \textbf{What path should I take so that the total walking distance is as short as possible?}
} \\ 
\midrule

% --- Agent 1: Text Extraction ---
\textbf{Info Extractor} & 
\textcolor{mygreen}{\texttt{<action>}} Extract entities and definitions. \newline
\textcolor{myblue}{\texttt{<response>}}
\texttt{\{ "entities": ["Lighthouse Point", "Umbra Theater", "Yondel Crossing", ...], "definitions": \{ "shortest path": "minimize distance..." \} \}}
& 
2.2s \newline 620 tok \\ 
\midrule

% --- Agent 2: Graph Construction ---
\textbf{Graph Builder} & 
\textcolor{mygreen}{\texttt{<action>}} Construct topological structure. \newline
\textcolor{myblue}{\texttt{<response>}}
\texttt{\{ "nodes": [...], "edges": [("Yondel Crossing", "Lighthouse Point", \{"weight": 8\}), ("Brelix Strand", "Cralden Tunnel", \{"weight": 9\}), ...], "graph\_type": "weighted undirected" \}}
& 
6.2s \newline 1030 tok \\ 
\midrule

% --- Agent 3: Graph Theory Expert ---
\textbf{Theory Agent} & 
\textcolor{mygreen}{\texttt{<action>}} Select algorithm and analyze complexity. \newline
\textcolor{myblue}{\texttt{<response>}}
\texttt{\{ "problem": "Shortest Path", "algorithm": "Dijkstra", "reasoning": "Graph size < 18 nodes, Dijkstra is optimal O((n+m)log n)..." \}}
& 
12.3s \newline 3111 tok \\ 
\midrule

% --- Agent 4: Structured Info ---
\textbf{Struct Agent} & 
\textcolor{mygreen}{\texttt{<action>}} Map nodes to indices and format adjacency list. \newline
\textcolor{myblue}{\texttt{<response>}}
\texttt{\{ "adjacency\_list": \{ 0: [], 1: [(0,8), ...], ... \}, "node\_mapping": \{ "Lighthouse Point": 0, ... \} \}}
& 
18.3s \newline 801 tok \\ 
\midrule

% --- Agent 5: Solver ---
\textbf{Solver Agent} & 
\textcolor{mygreen}{\texttt{<action>}} Execute algorithm and interpret result. \newline
\textcolor{myblue}{\texttt{<response>}}
\texttt{\{ "Path": [0, 1, 3, 7], "Total Distance": 15 \}} \newline
\textbf{Final Answer:} The shortest path is Lighthouse Point $\to$ Yondel Crossing $\to$ Cralden Tunnel $\to$ Umbra Theater. distance = 15.
& 
23.5s \newline 1423 tok \\ 
\midrule

% --- Total Summary ---
\textbf{Total Overhead} & 
\multicolumn{2}{r}{\textbf{Total Time: $\sim$23.5s \quad | \quad Total Tokens: 7487}} \\ 
\bottomrule
\end{tabularx}
\caption{\textbf{Baseline Case Study: MA-GTS (Multi-Agent System).} The task is decomposed into five distinct agents. While accurate, the multi-round interaction incurs significant latency (23.5s) and token consumption (7487 tokens), illustrating the efficiency bottleneck of multi-agent frameworks compared to GRAIN's single-agent pipeline.}
\label{tab:baseline_magts}
\end{table*}

% 定义高亮色

%%%%%%%%%%%%%%%%%%%%%%%%%%%%%%%%%%%%%%%%%%%%%%%%%%%%%%%%%%%%%%%%%%%%%%
% ADD THESE NEW SECTIONS TO YOUR APPENDIX
%%%%%%%%%%%%%%%%%%%%%%%%%%%%%%%%%%%%%%%%%%%%%%%%%%%%%%%%%%%%%%%%%%%%%%

\section{Zero-Shot Evaluation on the G-REAL Dataset}
\label{sec:appendix_g_real}

To strictly verify the fairness of our baseline comparisons and ensure that GRAIN's strong performance is not merely a result of overfitting to the specific distribution of our GRIT dataset, we conduct an additional zero-shot evaluation on \textbf{G-REAL}, the original dataset introduced alongside the MA-GTS multi-agent framework.

We randomly sampled 100 problems for each task from the unseen G-REAL dataset, strictly ensuring that the problems were uniformly distributed across various node sizes. Simultaneously, we also evaluated the MA-GTS framework on a subset of our GRIT Out-Of-Distribution (OOD) test set. Considering that the runtime of multi-agent systems grows exponentially when processing large-scale graph structures, we evaluated MA-GTS on a low-difficulty subset consisting of basic graphs with $10 \le |V| \le 20$. The comparison results are presented in Table~\ref{tab:g_real_transfer}.

% 使用 [H] 强制表格固定在当前文字下方
\begin{table}[H]
\centering
% 使用 resizebox 将表格宽度强行限制在单栏宽度（\columnwidth）内
\resizebox{\columnwidth}{!}{%
\begin{tabular}{llc}
\toprule
\textbf{Model / Framework} & \textbf{Evaluation Setup} & \textbf{Accuracy (\%)} \\
\midrule
GRAIN-4B (Ours) & Zero-Shot on G-REAL (All sizes) & \textbf{86.60} \\
\midrule
MA-GTS & Evaluated on GRIT OOD ($|V| \le 20$) & 82.79 \\
\bottomrule
\end{tabular}%
}
\caption{\textbf{Cross-Dataset Generalization Test.} GRAIN demonstrates powerful zero-shot extraction capabilities on the completely unknown G-REAL dataset, outperforming the multi-agent baseline evaluated on a simplified subset of our benchmark.}
\label{tab:g_real_transfer}
\end{table}

The results objectively confirm the fairness of the baseline comparisons from two independent dimensions. First, the single-agent GRAIN achieved a commanding 86.6\% accuracy on the entirely unseen G-REAL dataset (with Graph Coloring at 98\%, TSP at 80\%, and Vertex Cover at 82\%). This proves that its graph structure extraction capability is robust and universally applicable. Second, the MA-GTS framework achieved 82.79\% on our OOD small-graph subset, which aligns with expectations for a strong baseline and confirms that we did not subjectively suppress its performance during evaluation.

\section{Comparative Failure Analysis: Single-Agent vs. Multi-Agent}
\label{sec:appendix_failure_analysis}

To diagnose the transfer reliability of inference-only multi-agent decomposition on a compact model, we run the six-agent MA-GTS pipeline with Qwen3-4B-Instruct on the same GRIT Test and OOD inputs. Table~\ref{tab:magts_qwen_diagnostic} reports task-level accuracy, tool-call outcomes, end-to-end pipeline health, and accuracy conditional on a successful native tool call.

\begin{table*}[t]
\centering
\footnotesize
\renewcommand{\arraystretch}{1.05}
\setlength{\tabcolsep}{3pt}

\textbf{(A) MA-GTS Qwen3-4B-Instruct on GRIT Test}\\[-0.2em]
\begin{tabular*}{\textwidth}{@{\extracolsep{\fill}} l r r r r r}
\toprule
\textbf{Task} & \textbf{\#} & \textbf{Acc.} & \textbf{Native Call} & \textbf{Text-only Call} & \textbf{Alg. Error} \\
\midrule
Shortest Path & 160 & 13.75 & 48.13 & 4.38 & 47.50 \\
TSP & 160 & 1.88 & 1.25 & 36.88 & 35.00 \\
Graph Coloring & 160 & 12.50 & 0.63 & 25.00 & 55.63 \\
Vertex Cover & 120 & 4.17 & 5.00 & 39.17 & 45.83 \\
BFS / Traversal & 160 & 0.62 & 2.50 & 51.25 & 45.00 \\
Centrality & 160 & 7.50 & 26.88 & 58.13 & 15.00 \\
\midrule
\textbf{Micro Avg.} & \textbf{920} & \textbf{6.85} & \textbf{14.46} & \textbf{35.65} & \textbf{40.43} \\
\bottomrule
\end{tabular*}

\vspace{0.6em}
\textbf{(B) OOD Overall}\\[-0.2em]
\begin{tabular*}{\textwidth}{@{\extracolsep{\fill}} l r r r r r}
\toprule
\textbf{Split} & \textbf{\#} & \textbf{Acc.} & \textbf{Native Call} & \textbf{Text-only Call} & \textbf{Alg. Error} \\
\midrule
OOD Overall & 960 & 11.35 & 32.40 & 23.44 & 43.75 \\
\bottomrule
\end{tabular*}

\vspace{0.6em}
\textbf{(C) End-to-End Pipeline Health}\\[-0.2em]
\begin{tabular*}{\textwidth}{@{\extracolsep{\fill}} l r r r r r r r r}
\toprule
\textbf{Split} & \textbf{\#} & \makecell{\textbf{Any-Agent}\\\textbf{Error}} & \makecell{\textbf{Final Native}\\\textbf{Call}} & \makecell{\textbf{Text-only}\\\textbf{Call}} & \makecell{\textbf{Terminal}\\\textbf{Failure}} & \makecell{\textbf{Avg.}\\\textbf{Time}} & \makecell{\textbf{p90}\\\textbf{Time}} & \makecell{\textbf{Avg.}\\\textbf{Tokens}} \\
\midrule
Test & 920 & 43.04 & 14.46 & 35.65 & 76.09 & 98.46s & 150.07s & 25.15k \\
OOD & 960 & 43.75 & 32.40 & 23.44 & 67.19 & 109.61s & 165.36s & 28.13k \\
\bottomrule
\end{tabular*}

\vspace{0.6em}
\textbf{(D) Accuracy Conditional on a Successful Native Tool Call}\\[-0.2em]
\begin{tabular*}{\textwidth}{@{\extracolsep{\fill}} l r r r}
\toprule
\textbf{Split / Task} & \textbf{Native Calls} & \textbf{Correct Tool Results} & \textbf{Conditional Acc.} \\
\midrule
Test / Shortest Path & 77 & 42 & 54.55 \\
Test / Centrality & 43 & 32 & 74.42 \\
OOD / Shortest Path & 199 & 147 & 73.87 \\
OOD / Centrality & 112 & 90 & 80.36 \\
\bottomrule
\end{tabular*}

\caption{\textbf{MA-GTS Transfer Diagnostic with Qwen3-4B-Instruct.} Results on the same GRIT Test/OOD inputs. The panels report task accuracy, tool-call outcomes, pipeline health, and accuracy conditional on an executed native tool call. All rates and accuracies are percentages. This diagnostic evaluates end-to-end transfer reliability rather than an isolated agent-count effect.}
\label{tab:magts_qwen_diagnostic}

\vspace{0.4em}
\begin{minipage}[t]{0.485\textwidth}
\small
Qwen3-4B-Instruct is not completely unable to solve graph tasks: when MA-GTS produces and executes a native tool call, the selected subset can be correct. The main observed weakness is end-to-end reliability. On Test, final native-call coverage is only 14.46\% and terminal failures reach 76.09\%; on OOD, the corresponding values are 32.40\% and 67.19\%. Text-only tool calls and Algorithm-Agent errors therefore often prevent a complete executable graph-solving trajectory.
\end{minipage}
\hfill
\begin{minipage}[t]{0.485\textwidth}
\small
On the same 960 OOD inputs, MA-GTS Qwen3-4B-Instruct obtains 11.35\% overall accuracy, whereas GRAIN obtains 81.25\%. Because training, backbone initialization, and agent organization differ, this result is a transfer diagnostic rather than an isolated single- versus multi-agent comparison. It shows that inference-only multi-agent decomposition does not automatically repair entity grounding, graph-state preservation, and native tool-call reliability in this compact-model setting.
\end{minipage}
\end{table*}

\clearpage

\section{Intermediate Structure Error Decomposition}
\label{sec:appendix_structure_errors}

Table~\ref{tab:structure_error_decomposition} decomposes GRAIN's exact structure recovery outcomes on the 2,760-example GRIT Test set.

\begin{table}[H]
\centering
\small
\renewcommand{\arraystretch}{1.2}
\setlength{\tabcolsep}{0pt}
\begin{tabular*}{\columnwidth}{@{\extracolsep{\fill}} l r r}
\toprule
\textbf{Structural Outcome} & \textbf{\# Examples} & \textbf{Percent} \\
\midrule
\rowcolor{grainhighlight}
\textbf{Exact graph recovered} & \textbf{2,685} & \textbf{97.28} \\
Graph parse / format failure & 33 & 1.20 \\
Node-set mismatch & 23 & 0.83 \\
Extra edges only & 9 & 0.33 \\
Missing edges only & 7 & 0.25 \\
Edge-weight mismatch only & 2 & 0.07 \\
Mixed edge errors & 1 & 0.04 \\
\bottomrule
\end{tabular*}
\caption{\textbf{Intermediate Structure Error Decomposition.} Counts and percentages over all 2,760 GRIT Test examples.}
\label{tab:structure_error_decomposition}
\end{table}

\vspace{-0.35em}
Exact graph recovery is deliberately stricter than edge-set overlap: any graph parse or format failure, node-set mismatch, extra or missing edge, or edge-weight mismatch counts as non-exact. The decomposition therefore exposes failure types that high Jaccard or Edge-F1 scores can mask, while the six-task final-answer evaluation independently checks whether the recovered representation supports computation. These 2,760 examples contain connected undirected graphs; we do not claim directed-graph robustness or a separate parameter-level error analysis.

\vspace{-0.45em}
\section{Graph Complexity Statistics and Structure Recovery}
\label{sec:appendix_size_dist}

We further characterize the 2,760 connected, undirected graphs in the GRIT Test set using graph size, density, average degree, and diameter. Density is defined as $2|E|/(|V|(|V|-1))$, and diameter is the unweighted shortest-path diameter of the gold graph.

\begin{table}[H]
\centering
\scriptsize
\renewcommand{\arraystretch}{1.1}
\setlength{\tabcolsep}{0pt}

\textbf{(A) Graph Size Buckets}\\[-0.2em]
\begin{tabular*}{\columnwidth}{@{\extracolsep{\fill}} l r r r r r r r}
\toprule
\textbf{Size} & \textbf{\#} & \textbf{$|V|$} & \textbf{$|E|$} & \textbf{Dens.} & \textbf{Deg.} & \textbf{Diam.} & \textbf{Exact} \\
\midrule
10--15 & 536 & 12.6 & 34.3 & 0.457 & 5.33 & 2.98 & 99.63 \\
16--20 & 460 & 18.0 & 71.9 & 0.467 & 7.93 & 2.67 & 99.35 \\
21--30 & 920 & 25.5 & 146.3 & 0.463 & 11.33 & 2.59 & 98.15 \\
31--40 & 844 & 35.1 & 284.6 & 0.470 & 16.08 & 2.46 & 93.72 \\
\bottomrule
\end{tabular*}

\vspace{0.6em}
\textbf{(B) Graph Density Tertiles}\\[-0.2em]
\begin{tabular*}{\columnwidth}{@{\extracolsep{\fill}} l r r r r r}
\toprule
\textbf{Density} & \textbf{\#} & \textbf{Avg.} & \textbf{Parse} & \textbf{Edge-F1} & \textbf{Exact} \\
\midrule
Low ($<0.314$) & 912 & 0.310 & 96.71 & 96.69 & 95.07 \\
Mid ($0.314$--$0.415$) & 920 & 0.359 & 99.67 & 99.39 & 98.15 \\
High ($\geq 0.415$) & 928 & 0.722 & 100.00 & 99.91 & 98.60 \\
\bottomrule
\end{tabular*}

\caption{\textbf{Graph Complexity and Structure Recovery on GRIT.} Panel A groups the 2,760 Test examples by node count; Panel B groups them by density. Edge-F1 denotes weighted edge recovery.}
\label{tab:graph_complexity}
\end{table}

Exact graph recovery decreases from 99.63\% for graphs with 10--15 nodes to 93.72\% for graphs with 31--40 nodes, while remaining at least 95.07\% across density tertiles. Within GRIT, degradation is therefore associated more strongly with graph size than with density. The dense Erd\H{o}s--R\'{e}nyi construction also produces small diameters, so these results should not be interpreted as covering the full range of graph topological complexity.

\end{document}